\documentclass{article}

\usepackage{microtype}
\usepackage{graphicx}
\usepackage{subfigure}
\usepackage{booktabs} 

\usepackage{hyperref}
\usepackage{enumitem}
\setlist{nolistsep}

\usepackage[accepted]{icml2020}

\usepackage{amsfonts, mathrsfs, amssymb}       
\usepackage{graphicx}
\usepackage{subfigure,wrapfig}
\usepackage{dsfont}
\usepackage{amsmath}
\usepackage{amsthm}
\usepackage{stmaryrd}
\usepackage{xcolor}
\usepackage{nicefrac}       

\usepackage{etoolbox}
\usepackage{subfiles}
\usepackage{algorithm}
\usepackage{algorithmic}
\usepackage{shortcuts}
\usepackage{wrapfig}

\newtoggle{supplementary}
\toggletrue{supplementary}

\icmltitlerunning{Debiased Sinkhorn barycenters}

\begin{document}

\twocolumn[
\icmltitle{Debiased Sinkhorn barycenters}

\begin{icmlauthorlist}
	\icmlauthor{Hicham Janati}{a,b}
	\icmlauthor{Marco Cuturi}{c,b}
	\icmlauthor{Alexandre Gramfort}{b}
\end{icmlauthorlist}

\icmlaffiliation{b}{CREST-ENSAE, France}
\icmlaffiliation{a}{Inria Saclay, France}
\icmlaffiliation{c}{Google Research, Brain team, France}
\icmlcorrespondingauthor{Hicham Janati}{hicham.janati@inria.fr}


\vskip 0.3in
]



\printAffiliationsAndNotice{}  

\begin{abstract}
Entropy regularization in optimal transport (OT) has been the driver of many recent interests for Wasserstein metrics and barycenters in machine learning. It allows to keep the appealing geometrical properties of the unregularized Wasserstein distance while having a significantly lower complexity thanks to Sinkhorn's algorithm. However, entropy brings some inherent \emph{smoothing bias}, resulting for example in blurred barycenters. This side effect has prompted an increasing temptation in the community to settle for a slower algorithm such as log-domain stabilized Sinkhorn which breaks the parallel structure that can be leveraged on GPUs, or even go back to unregularized OT. Here we show how this bias is tightly linked to the reference measure that defines the entropy regularizer and propose debiased Wasserstein barycenters that preserve the best of both worlds: fast Sinkhorn-like iterations without entropy smoothing. Theoretically, we prove that the entropic OT barycenter of univariate Gaussians is a Gaussian and quantify its variance bias. This result is obtained by extending  the differentiability and convexity of entropic OT to sub-Gaussian measures with unbounded supports.  Empirically, we illustrate the reduced blurring and the computational advantage on various applications.
\end{abstract}

\section{Introduction}
\label{s:intro}
Comparing, interpolating or averaging probability distributions is an ubiquitous problem in machine learning. Optimal transport (OT) offers an efficient way to do exactly that while taking into account the geometry of the space they live in~\cite{otbook}. Let $\cP(\bbR^d)$ denote the set of probability measures on $\bbR^d$. Given some divergence $F: \cP(\bbR^{d} )\times \cP(\bbR^{d} ) \to \bbR$ and weights $(w_k)_k$ such that $\sum_{k=1}^K w_k = 1$, the weighted barycenter of a set of probability measures $(\alpha_k)_k$ can be defined as the Fr\'echet mean:
\vspace{-0.3cm}
\begin{equation}
\label{eq:frechet}
\alpha_{F}\eqdef~{\argmin}_{\alpha \in \cP({\bbR^d})} \sum_{k=1}^K w_k F(\alpha_k, \alpha)\enspace.
\end{equation}
Here $\alpha_{F}$ can be thought as a weighted average of distributions. While the $(\alpha_k)_k$ may have
a fixed support or known finite supports when working in machine learning applications, the support of
$\alpha_{F}$ may or may not be known.
When the latter is unknown a priori, \emph{free support methods} are needed to jointly minimize the objective with respect to both the support and the mass of the distribution~\citep{cuturi14}. Otherwise, \emph{fixed support methods}, which only optimize weights on known supports, are employed~\citep{benamou14}. While free support methods are more general and memory efficient, fixed support ones are faster in practice. In this paper, we focus on fixed support methods.
 
Using the Wasserstein distance as a divergence $F$,~\citet{li08} were the first to propose the Fr\'echet mean \eqref{eq:frechet} for a clustering application in computer vision. This idea was later adopted by~\citet{agueh:11} to formally define Optimal Transport (OT) barycenters. However, the Wasserstein distance is defined through a linear programming problem which does not scale to large datasets. To address this computational issue, some form of regularization is mandatory: either regularize the measures themselves using sliced projections for instances or regularize the OT problem using $\ell_2$~\citep{blondel18a} or entropy~\citep{cuturi13}. While $\ell_2$ preserves some of the sparsity of the non-regularized optimal transportation plan, entropy regularization leads to an approximation of the Wasserstein distance that can be solved using a fast and parallelizable GPU-friendly algorithm: the celebrated Sinkhorn's algorithm~\citep{cuturi13}. In the rest of this paper, we will focus on entropic OT.
Let $\cost$ be a non-negative  cost function on $\bbR^d \times \bbR^d$ such that $\cost(x, y) = 0 \Leftrightarrow x = y$. For instance,  a usual choice is $\cost(x, y) = \|x - y\|^2$. Entropy regularized OT between $\alpha, \beta \in \cP(\bbR^d)$ with the reference measures $m_1, m_2 \in \cP({\bbR^d})$ is defined as:
 \begin{align}
 \label{eq:otdef}
 \begin{split}
 \ote^{m_1, m_2}(\alpha, \beta )\eqdef & \\
 \min_{\substack{\pi \in \cP({\bbR^d \times \bbR^d}) \\ \pi_{\#1} = \alpha, \pi_{\#2} = \beta}} \int_{\bbR^{d \times d}} \cost \diff\pi & + \varepsilon \kl(\pi | m_1 \otimes m_2) \enspace,
 \end{split}
 \end{align}
 where $\varepsilon > 0$, $\pi_{\#1}, \pi_{\#2}$ denote the left and right marginals of $\pi$ respectively, $m_1 \otimes m_2$ is the product measure of $m_1$ and $m_2$, and the relative entropy is defined as: 
 \begin{equation}
 \label{eq:kl}
 \kl(\pi | m_1\otimes m_2) \eqdef    \int_{\bbR^d \times \bbR^d} \log\left(\frac{\diff\pi}{\diff(m_1 \otimes m_2)}\right)\diff\pi \enspace .
 \end{equation}
 Naturally, in the discrete case, \citet{benamou14} proposed to compute OT barycenters of discrete measures using $F= \ote^{m_1, m_2}$ with $m_1 = m_2 = \cU$, the uniform measure over the finite set on which the measures are defined. Doing so, they showed that the barycenter problem is equivalent to Iterative Bregman Projections (IBP) which are similar to Sinkhorn's scaling operations. However, entropy regularization leads to an undesirable blurring of the barycenter. While using a very small regularization may appear as an obvious solution, it leads to numerical instabilities that can only be mitigated using log-domain stabilization or full log-domain `logsumexp' operations \citep{schmitzer16}. This however considerably slows down Sinkhorn's iterations.
 
 To reduce this entropy bias, several divergences $F$ have been proposed.  For instance, \citet{solomon15} proposed to modify the IBP algorithm by adding a maximum entropy constraint they called \emph{entropy sharpening}. This leads to a non-convex constraint which does not fit within the IBP framework. \citet{luise18} proposed to compute the entropy regularized solution $\pi^\star$ and to evaluate the OT loss \eqref{eq:otdef} without the entropy term $\kl$. This indeed leads to sharper barycenters but can only be estimated via gradient descent, thus requiring a full Sinkhorn loop at each iteration and setting a pre-defined learning rate which can be cumbersome in practice. \citet{amari19} proposed a modified entropy regularized divergence OT that can still leverage the fast IBP algorithm of \citet{benamou14} but requires a final deconvolution step with the kernel $\exp(-\frac{\cost}{\varepsilon})$, which is only feasible when $\varepsilon$ is small. With this same objective of non-blurred solutions, \citet{dongdong19} even called for a return to the original non-regularized Wasserstein barycenter and proposed an accelerated interior point methods algorithm.
 
 \paragraph{Our main contributions}
 Except \cite{dongdong19}, all the works proposed above employ the uniform measure as reference, i.e they use $\ote^\cU \eqdef \ote^{m_1, m_2}$ with $m_1 = m_2 = \cU$. The purpose of this paper is to highlight a direct link between the already known entropy bias of the OT barycenter and this particular choice of $m_1$ and $m_2$. This link is illustrated by showing how the choice of $m_1$ and $m_2$ impact the barycenter of univariate Gaussians in $\bbR^d$. Following \citep{ramdas17,genevay18,feydy19, luise19}, we advocate for using the following Sinkhorn divergence which can be defined without specifying $m_1$ and $m_2$ for arbitrary measures $\alpha, \beta \in \cP(\bbR^d)$:
 \begin{equation*}
 \label{eq:sinkhorn-div-def}
 \sdiv(\alpha, \beta) \eqdef  \ote(\alpha, \beta) - \frac{\ote(\alpha, \alpha) + \ote(\beta, \beta)}{2}\enspace.
 \end{equation*}
The choice of the reference measures $m_1$ and $m_2$ has led to different formulations of regularized OT. The main contributions of this paper are twofold. (1) theoretical: we quantify the entropy bias of usual reference measures for univariate Gaussians. Precisely, while the Lebesgue measure ($m_1=m_2 = \cL$) induces a blurring bias and the product measure ($m_1=\alpha, m_2=\beta$) induces a shrinking bias, $\sdiv$ is actually debiased. (2) empirical: we propose a fast iterative algorithm similar to IBP to compute debiased barycenters. Unlike other gradient-based methods, this fixed point algorithm can be efficiently differentiated with respect to the barycentric weights via backpropagation. This allows one to carry out Wasserstein barycentric projections without entropy blurring. This will be illustrated in the experiments.

In the following section we discuss the different choices of $m_1$ and $m_2$ and quantify their induced entropy bias upon the barycenters of univariate Gaussians. In Section \ref{s:gaussians}, we show some useful properties of $\sdiv$ (differentiability, convexity) when defined on sub-Gaussian measures with unbounded supports in $\bbR^d$ which are necessary to prove the theorems of section \ref{s:reference}. Next, in Section \ref{s:algorithm} we turn to computational aspects and provide a fast Sinkhorn-like algorithm for debiased barycenters. We conclude with numerical experiments in Section \ref{s:experiments}.
%

\section{Reference measure and entropy bias}
\label{s:reference}

\paragraph{Notation}
We denote by $\mathds 1$ the vector of ones in $\bbR^n$. On matrices, $\log$, $\exp$ and the division operator are applied element-wise. We use $\odot$ for the element-wise multiplication between matrices or vectors.  On vectors and matrices, the same notation denotes the usual scalar products: for $x, y \in \bbR^n$, $\langle x, y\rangle = \sum_{i=1}^n x_i y_i$; and for matrices $\bA, \bB \in \bbR^{n, n}$,  $\langle \bA, \bB\rangle = \sum_{i,j=1}^n \bA_{ij} \bB_{ij}$. 

\paragraph{Uniform reference and IBP}
Let $\cX = \{x_1, \dots, x_n\} \subset \bbR^d$ and consider two discrete measures $\alpha = \sum_{i=1}^n\alpha_i \delta_{x_i}$ and $\beta = \sum_{i=1}^n\beta_i \delta_{x_i}$. One can identify $\alpha$ and $\beta$ with their weights $\alpha_i$ and $\beta_i$ where $\alpha^\top \mathds 1 = \beta^\top \mathds 1 $. Let $\bC \in \bbR_+^{n\times n}$ be the matrix such that $\bC_{ij} = \cost(x_i, x_j)$. The definition of $\ote^{\cU}$ in \eqref{eq:otdef} becomes:
\begin{equation}
\label{eq:otdef-uniform-discrete}
\ote^{\cU}(\alpha, \beta) = 
\min_{\substack{\pi \in \bbR^{n \times n}_+ \\ \pi \mathds 1 = \alpha, \pi^\top \mathds 1 = \beta}} \langle \bC, \pi\rangle+ \varepsilon \kl(\pi|\bU)\enspace,
\end{equation}
where $\bU$ is the uniform measure on $\cX^2$ given by $\frac{\mathds 1\mathds 1^\top}{n^2}$.
Let $\bK$ be the element-wise exponentiated kernel $\exp\left(-\frac{\bC}{\varepsilon}\right)$. By adopting the definition $\widetilde{\kl}(\bA, \bB) = \sum_{i,j}^n\bA_{ij} \log\left(\frac{\bA_{ij}}{\bB_{ij}}\right) + \bB_{ij} - \bA_{ij}$ for $\bA, \bB \in \bbR_+^{n\times n}$, \citet{benamou14} noticed that \eqref{eq:otdef-uniform-discrete} is equivalent to a Kullback-Leibler projection up to an additive constant:
\begin{equation}
\label{eq:otdef-uniform-kl}
\ote^{\cU}(\alpha, \beta)=
\min_{\substack{\pi \in \bbR^{n \times n}_+ \\ \pi \mathds 1 = \alpha, \pi^\top \mathds 1 = \beta}}  \varepsilon \widetilde{\kl}(\pi | \bK)
\end{equation}
and proposed the Iterative Bregman Projections (IBP) algorithm to solve the equivalent barycenter problem:
\begin{equation}
\label{eq:ibp}
\min_{\substack{\pi_1, \dots, \pi_K \\ \pi_k \in \cC_k \cap \cC'}} \sum_{k=1}^K w_k \widetilde{\kl}(\pi^k| \bK)\enspace,
\end{equation}
where $\cC_k = \{\pi \in \bbR^{n\times n}_+ | \pi \mathds 1 = \alpha_k\}$ and $\cC'= \{\pi \in \bbR^{n\times n}_+ | \exists \alpha \in \Delta_n, \enspace \pi_k^\top \mathds 1 = \alpha, \enspace \forall k = 1\dots K\}$.
The IBP algorithm amounts to performing iterative minimization on one constraint set at a time. Each step can be solved in closed form, leading to Sinkhorn-like iterations, see supplementary section \ref{ss:ibp} for details on IBP.

\begin{figure}[t!]
	\centering
		\includegraphics[width=0.9\linewidth, trim={3cm 0.7cm 0 0.5cm}, clip]{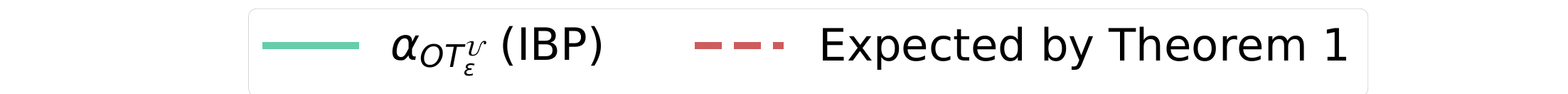}
	
	\includegraphics[width=\linewidth]{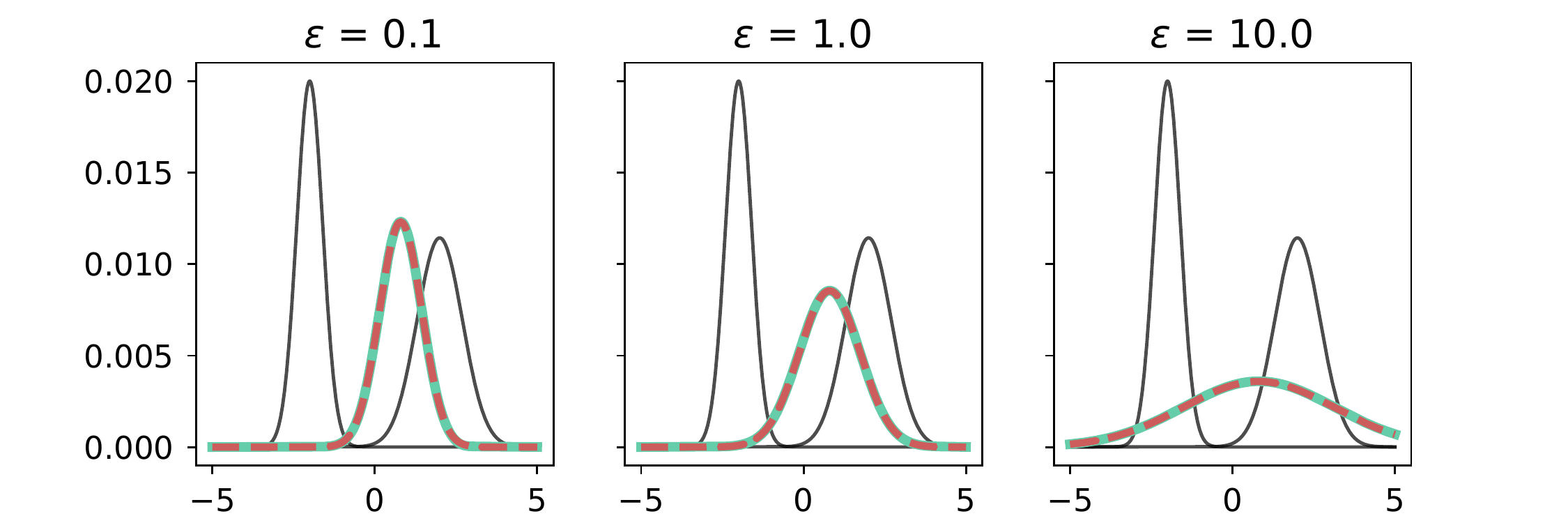}
	\vspace{-0.6cm}
	\caption{Illustration of theorem \ref{thm:lebesgue} with $\cN(-2, 0.4)$ and $\cN(2, 0.7)$ shown in black, and $(w_1, w_2) = (0.4, 0.6)$. The barycenter $\ote^{\cU}$ matches theoretical expectations and is biased towards blurred distributions.
	\label{f:gaussians-blur}}
\end{figure}

\paragraph{Lebesgue reference and smoothing bias}
As discussed in the introduction, the obtained barycenter $\alpha_{\ote^\cU}$ suffers from entropy blurring. To quantify this blur, we turn to Lebesgue continuous measures and consider the Lebesgue measure as a reference by setting $m_1 = m_2 = \cL$. We argue that by considering normalized histograms, the discrete formulation \eqref{eq:otdef-uniform-kl} provides an approximation of $\ote^{\cL}$ when the number of histogram bins tends to $+\infty$. Indeed, since $\ote^{\cL}$ is defined on Lebesgue-continuous measures, one can identify $\alpha, \beta$ and $\pi$ with their density functions. Moreover, if the density functions are positive, the same $\kl$ factorization \eqref{eq:otdef-uniform-kl} is possible for $\ote^{\cL}$.  The following theorem shows that the weighted barycenter of univariate Gaussians is Gaussian with an increased variance. Figure~\ref{f:gaussians-blur} illustrates this smoothing bias using discrete histograms with a grid of 500 bins.
\begin{theorem}[Blurring bias of $\ote^{\cL}$]
	\label{thm:lebesgue}
	Let $\cost(x, y) = (x - y)^2$ $\varepsilon >0$ and $\varepsilon = 2\varepsilon'^2$. Let $(w_k)_k$ be positive weights that sum to 1. Let $\cN$ denote the Gaussian distribution. Assume $\alpha_k \sim \cN(\mu_k, \sigma_k^2)$  and let $\bar{\mu} = \sum_k w_k \mu_k$,
	
	then:
	
	(i) $\alpha_{\ote^{\cL}}  \sim \cN(\bar{\mu}, S^2) $ where $S$ is a positive solution of the equation: 
	$\sum_{k=1} w_k \sqrt{\varepsilon'^4 + 4\sigma_k^2S^2} = - \varepsilon'^2 + 2S^2$. 

	(ii) In particular, if all $\sigma_k$ are equal to some $\sigma > 0$,
	
	then $\alpha_{\ote^{\cL}} \sim \cN(\bar{\mu}, \sigma^2 + \varepsilon'^2) $.
\end{theorem}
\proof{See section \ref{ss:proof-lebesgue}}

\begin{figure}[t!]
	\centering
	\includegraphics[width=0.9\linewidth, trim={2cm 0.7cm 0 0.5cm}, clip]{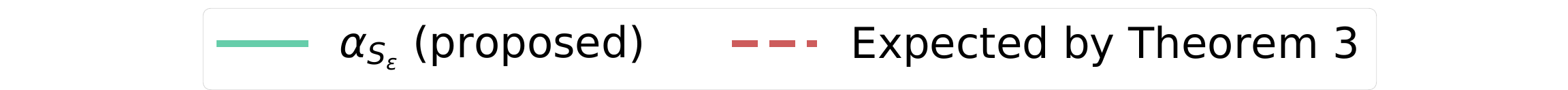}
	
	\includegraphics[width=\linewidth]{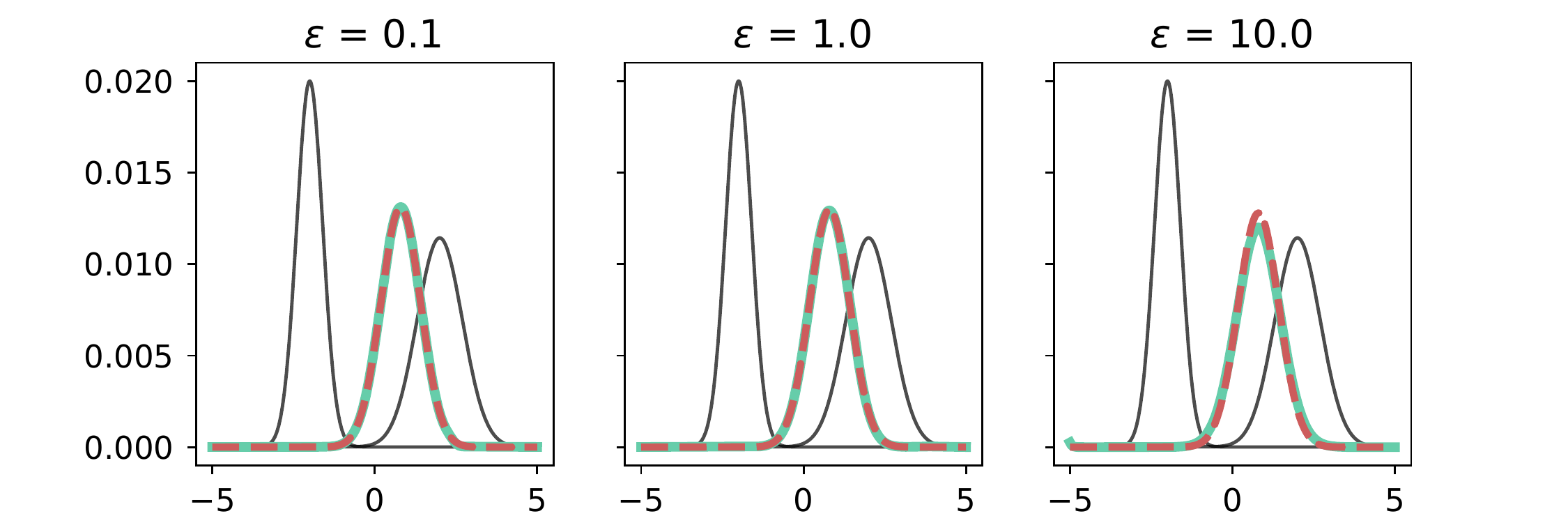}
	\vspace{-0.5cm}
	\caption{Illustration of theorem \ref{thm:div}. Unlike with the uniform measure (Figure \ref{f:gaussians-blur}), the debiased barycenter remains unscathed when increasing~$\varepsilon$. \label{f:gaussians-div}}
\end{figure}

\begin{figure}[t!]
	\centering
		\includegraphics[width=0.9\linewidth, trim={2cm 0.7cm 0 0.5cm}, clip]{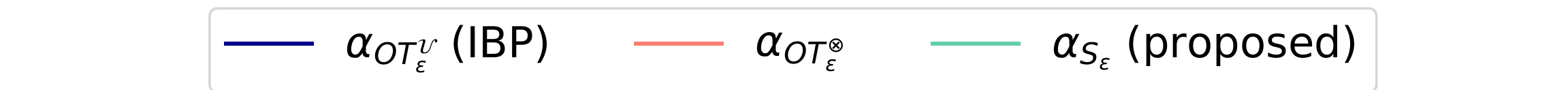}
	
	\includegraphics[width=\linewidth]{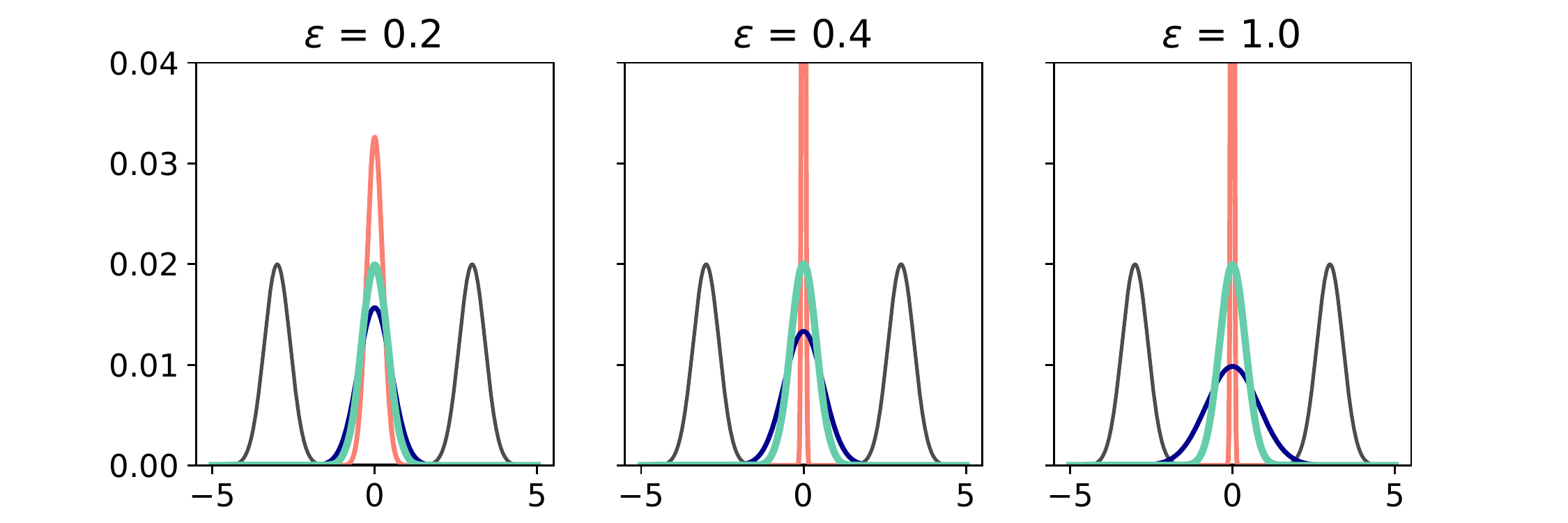}
	\vspace{-0.5cm}
	\caption{Illustration of the three theorems  with $\cN(-3, 0.4)$ and $\cN(3, 0.4)$ shown in black using uniform weights. Entropy regularization causes a smoothing bias (blue) and a shrinking bias (red). Debiasing with $\sdiv$ (cyan) is perfect and independent of $\varepsilon$.  \label{f:gaussians}}
\end{figure}

\paragraph{The product measure and shrinking bias}
Besides the smoothing bias of the uniform measure, $\ote^\cU$ cannot be generalized to a general OT definition for any arbitrary distributions that are non-discrete or non-Lebesgue continuous measures. To go beyond this binary classification of probability measures, several authors \citep{ramdas17, genevay18, feydy19} proposed the generic references $m_1 = \alpha$, $m_2 = \beta$. Indeed, the marginal constraints $\pi_1 = \alpha$, $\pi_2 = \beta$ imply that the support of $\pi$ is included in that of $\alpha \otimes \beta$ and the $\kl$ term is always well-defined regardless of the nature of $\alpha$ and $\beta$. For the sake of convenience, we denote $\ote^\otimes \eqdef \ote^{\alpha, \beta}$. \citet{dimarino19} made the following key observation that characterizes the change of reference.  For discrete measures $\alpha, \beta$:
\begin{equation}
\label{eq:reference-u}
\ote^\cU(\alpha, \beta) = \ote^\otimes(\alpha, \beta) 
+\varepsilon \kl(\alpha| \cU) + \varepsilon \kl(\beta|\cU)\enspace.
\end{equation}
Similarly, the same identity holds for Lebesgue-continuous measures in $\cP(\bbR^d)$:
\begin{equation}
\label{eq:reference-l}
\ote^\cL(\alpha, \beta) = \ote^\otimes(\alpha, \beta) 
+\varepsilon \kl(\alpha| \cL) + \varepsilon \kl(\beta|\cL)\enspace.
\end{equation}
The identity \eqref{eq:reference-u} unveils another merit of $\ote^{\otimes}$ over $\ote^\cU$: its corresponding barycenter problem is equivalent to a regularized $\ote^\cU$ barycenter with a negative $\kl$ penalty. Interestingly, even though `$-\kl$' is concave, $\ote^{\otimes}$ remains convex with respect to one of its arguments \citep{feydy19}. However, $\ote^{\otimes}$ yet suffers from some limitations: (1) $\ote^{\otimes}$ cannot be written as a KL projection, thus the fast IBP algorithm is lost; 
(2) the barycenter $\alpha_{\ote^{\otimes}}$ of Gaussians can be a degenerate Gaussian, as demonstrated by Theorem \ref{thm:product} which shows that if $\varepsilon$ is large, the barycenter collapses to a Dirac (cf. Figure \ref{f:gaussians}). This phenomenon can however be leveraged as a deconvolution technique: \citet{rigollet18} showed that minimizing $\ote^\otimes$ is equivalent to maximum-likelihood deconvolution of an additive Gaussian-noise model.
\begin{theorem}[Shrinking bias of $\ote^{\otimes}$]
	\label{thm:product}
	Let $\cost(x, y) = (x - y)^2$, $\varepsilon > 0$ and $\varepsilon = 2\varepsilon'^2$. Let $(w_k)_k$ be positive weights that sum to 1. Let $\cN$ denote the Gaussian distribution. Assume that $\alpha_k \sim \cN(\mu_k, \sigma_k^2)$  and let $\bar{\mu} = \sum_k w_k \mu_k$, $\bar{\sigma}^2 = \sum_{k=1}{w_k} \sigma_k^{2}$:
	
(i)	if $\varepsilon'^2 < \bar{\sigma}^2$ then $\alpha_{\ote^{\otimes}} \sim \cN(\bar{\mu}, S^2) $ where $S$ is a positive solution of the equation: $\sum_{k=1} w_k \sqrt{\varepsilon'^4 + 4\sigma_k^2S^2} = \varepsilon'^2 + 2S^2$. In particular, if all $\sigma_k$ are equal to some $\sigma > 0$, then
	$\alpha_{\ote^{\otimes}}~\sim~\cN(\bar{\mu}, \sigma^2 - \varepsilon'^2) $.
	
(ii) if $\varepsilon'^2 \geq \bar{\sigma}^2$ then $\alpha_{\ote^{\otimes}}$ is a Dirac  located at $\bar{\mu}$.
\end{theorem}
\proof{See section \ref{s:gaussians}.}
\paragraph{Debiased barycenters} 
Interestingly, these limitations and significant differences between $\ote^\cU$, $\ote^{\cL}$ and $\ote^{\otimes}$ disappear when considering the following Sinkhorn divergences:
\begin{equation*}
\label{eq:sinkhorn-div-unif}
\sdiv^m(\alpha, \beta) \eqdef  \ote^m(\alpha, \beta) - \frac{\ote^m(\alpha, \alpha) + \ote^m(\beta, \beta)}{2}\enspace,
\end{equation*}
\begin{equation*}
\label{eq:sinkhorn-div-def-prod}
\sdiv(\alpha, \beta) \eqdef  \ote^{\otimes}(\alpha, \beta) - \frac{\ote^{\otimes}(\alpha, \alpha) + \ote^{\otimes}(\beta, \beta)}{2}\enspace.
\end{equation*}
Using \eqref{eq:reference-u} and \eqref{eq:reference-l} it holds:
\begin{equation}
\label{eq:sinkhorn-div-ref}
\sdiv(\alpha, \beta) = \sdiv^m(\alpha, \beta)\enspace,
\end{equation}
where $m$ is either $\cU$ or $\cL$ depending on the nature of $\alpha$ and $\beta$. Therefore, $\sdiv$ is defined on arbitrary probability measures which can be mixtures of continuous measures and Dirac masses. Moreover, \citet{feydy19} showed that when the support of the measures is compact and with the additional assumption that $\cost$ is negative semi-definite, $\sdiv$ is differentiable and convex with respect to one of its arguments. In the following section, we generalize the aforementioned statements for measures with unbounded supports in $\bbR^d$. 
The negativity assumption on $\cost$ holds for instance if $C(x, y) = \|x - y\|^d$ with $0 < d \leq 2$~\citep[Chapter 3, Cor 3.3]{berg84} and is the only (cheap) price to pay for a debiased OT divergence. These convexity and differentiability results are essential to prove the debiasing of $\sdiv$ stated in Theorem \ref{thm:div} and illustrated in Figure~\ref{f:gaussians-div}.
\begin{theorem}[Debiasing  of $\sdiv$]
	\label{thm:div}
	Let $\cost(x, y) = (x - y)^2$ and $0 <~\varepsilon~<~+\infty$ and $\varepsilon = 2\varepsilon'^2$. Let $(w_k)_k$ be positive weights that sum to 1. Let $\cN$ denote the Gaussian distribution. Assume that $\alpha_k \sim \cN(\mu_k, \sigma_k^2)$  and let $\bar{\mu} = \sum_k w_k \mu_k$	then:
	
(i) $\alpha_{\sdiv} \sim \cN(\bar{\mu}, S^2) $ where $S$ is a positive solution $S^\star$ of the equation: 
	
$\sum_{k=1} w_k \sqrt{\varepsilon'^4 + 4\sigma_k^2S^2} = \sqrt{\varepsilon'^4 + 4S^4}$. Moreover, given a sorted sequence $\sigma_{(1)} \leq \dots \leq \sigma_{(K)}$, it holds $S^\star \in (\sigma_{(0)}, \sigma_{(K)})$.

(ii)	In particular, if all $\sigma_k$ are equal to some $\sigma > 0$, then $\alpha_{\sdiv}~\sim~\cN(\bar{\mu}, \sigma^2) $.
\end{theorem}
\proof{See section \ref{s:gaussians}.}

Figure \ref{f:gaussians} shows a comparison of the three barycenters discussed in this section. We intentionally chose Gaussians with equal variances to emphasize two observations: (1) the debiasing of $\sdiv$: the barycenter $\alpha_{\sdiv}$ has the same variance of the input measures for all $\varepsilon$; (2) the shrinking bias of $\ote^{\otimes}$ is significant even for small values of $\varepsilon$.

Besides debiasing, the barycenter $\alpha_{\sdiv}$ also comes with a computational advantage. Using the identity \eqref{eq:sinkhorn-div-ref}, we \emph{bypass} the technical difficulties of the product measure in $\sdiv$ and derive an algorithm similar to IBP to compute $\alpha_{\sdiv}$ which will be the subject of section \ref{s:algorithm}.
%
\section{$\sdiv$ is convex and differentiable on sub-Gaussian measures with unbounded supports}
\label{s:gaussians}

\paragraph{Notation}
The set of continuous function on $\bbR^d$ is denoted by $\cC(\bbR^d)$.
The set of probability measures with a second order moment is denoted by $\cP_2(\bbR^d)$.  For $\alpha\in \cP(\bbR^d)$, $\cL_p(\bbR^d, \alpha) $ denotes the set of continuous functions $\bbR^d \to \bbR$ such that $\int|f|^p\diff \alpha < +\infty$. Let $ f \in \cL_1(\bbR^d, \alpha)$, $g \in \cL_1(\bbR^d, \beta)$ and denote $\langle \alpha, f \rangle = \int_{\bbR^d} f\diff \alpha$. The tensor operators $\otimes$ and $\oplus$ denote respectively the mappings $f\otimes g: (x, y) \in \bbR^{d} \times \bbR^{d} \mapsto f(x).g(y)$ and $f\oplus g: (x, y) \in \bbR^{d} \times \bbR^{d} \mapsto f(x) + g(y)$.

To prove theorems \ref{thm:product}  and \ref{thm:div}, we characterize the optimality condition of the barycenter problem. First, we show that $\ote^{\otimes}$ and $\sdiv$ are convex (w.r.t. one variable) and differentiable. Our differentiability proof is inspired from that of \citet{feydy19} where the compactness assumption of the whole $\cX$ is replaced with a sub-Gaussian tails assumption on the measures that allows one to apply Lebesgue's dominated convergence theorem on $\bbR^d$. The convexity proof is however novel and is solely based on the dual problem of $\ote^{\otimes}$.
Proving theorem \ref{thm:lebesgue} requires studying $\ote^{\cL}$ which involves a slightly different dual problem. Since the differences are purely technical, we defer the proof of theorem \ref{thm:lebesgue} in the appendix and focus in this section on the product measure $\ote^\otimes$  and $\sdiv$ for the sake of clarity.
\paragraph{Dual problem}
In this section, we set  $\cost(x, y) = \|x - y\|^2$ with its associated Gaussian kernel $\kernel(x, y) = e^{-\frac{\|x -  y\|^2}{\varepsilon}}$. Let $\alpha, \beta \in  \cP(\bbR^d)$. 
 We define the linear operators on $\cK$ and $\cK^\top$ such that 
$\cK(\mu) = \int_{\bbR^d} \kernel(x, y) \diff \mu(y)$ and $\cK^\top(\mu) = \int_{\bbR^d} K^\top(x, y) \diff \mu(x)$ for any non-negative measure $\mu \in \cM_+(\bbR^d)$. 
 Problem \eqref{eq:otdef} has a dual formulation given by:
\begin{align}
\begin{split}
\label{eq:otdual}
\ote^\otimes(\alpha, \beta) = \sup_{\substack{f \in \cL_1(\bbR^d, \alpha) \\ g \in \cL_1(\bbR^d, \beta)}} \int_{\bbR^d} f\diff \alpha + \int_{\bbR^d} g\diff \beta  \\
-\varepsilon \int_{\bbR^{d} \times \bbR^{d}} \exp\left(\frac{f \oplus g - C}{\varepsilon}\right) \diff \alpha \diff \beta  + \varepsilon\enspace.
\end{split}
\end{align}
if $\alpha$ and $\beta$ have finite second moments, \eqref{eq:otdual} is well defined and a couple of dual potentials $(f, g)$ are optimal if and only if they are solutions of  Sinkhorn's equations \citep{mena19}:
\begin{align}
\begin{split}
\label{eq:schrodinger}
e^{\frac{f}{\varepsilon}}. \cK(e^{\frac{g}{\varepsilon}}. \beta) = 1, \enspace \alpha-a.e\enspace,\\
e^{\frac{g}{\varepsilon}}. \cK^\top(e^{\frac{f}{\varepsilon}}.\alpha) = 1, \enspace \beta-a.e\enspace.
\end{split}
\end{align}
and the optimal transport plan $\pi$ is given by:
$\pi~=~\exp\left(\frac{f \oplus g - C}{\varepsilon}\right).(\alpha \otimes \beta)$

Thus, at optimality the integral over $\bbR^{d} \times \bbR^{d}$ sums to 1 and:
\begin{equation}
\label{eq:ot-optimality}
\ote^{\otimes}(\alpha, \beta) = \int_{\bbR^d} f\diff \alpha + \int_{\bbR^d} g\diff \beta 
\end{equation}

\paragraph{Symmetric terms $\ote^{\otimes}(\alpha, \alpha)$}
When $\alpha = \beta$, the symmetry of the problem leads to the existence of a symmetric pair of potentials $(h, h)$. Indeed, if $(f, g)$ is optimal $(g, f)$ is also optimal. Moreover, since $C$ is symmetric, the optimal transport plan $\pi$ is also symmetric which leads to $f = g$. Thus the following proposition holds.

\begin{prop}
	\label{prop:symmetry}
	Let $\alpha \in \cP_2(\bbR^d)$, it holds:
	\begin{align}
		\label{eq:dual-sym}
		\begin{split}
		\ote^{\otimes}(\alpha, \alpha) = &\sup_{h\in \cL_1(\bbR^d, \alpha)} 2\int_{\bbR^d} h\diff \alpha  \\
		&-\varepsilon \int_{\bbR^{d} \times \bbR^{d}} \exp\left(\frac{h \oplus h - C}{\varepsilon}\right) \diff^2 \alpha + \varepsilon\enspace,
		\end{split}
	\end{align}
	Moreover, the supremum is attained at the unique (by strong concavity; since $C$ is definite negative)  autocorrelation potential $h\in \cL_1(\bbR^d, \alpha)$ if and only if $h$ is a solution of $e^{\frac{fh}{\varepsilon}}. \cK(e^{\frac{h}{\varepsilon}}. \alpha) = 1, \enspace \alpha-a.e\enspace$,
	and at optimality it holds: $\frac{1}{2}\ote^{\otimes}(\alpha, \alpha) = \int_{\bbR^d} h\diff\alpha$.
\end{prop}

\paragraph{Restriction on sub-Gaussians}
To derive theorems \ref{thm:product} and \ref{thm:div}, we show that both $\ote^{\otimes}$ and $\sdiv$ are convex and differentiable and provide a solution of the first order optimality condition. Notice that the convexity of $\ote^{\otimes}$ with respect to $\alpha$ and with respect to $\beta$ follows immediately from \eqref{eq:otdual} since it corresponds to a supremum of linear functionals. \citet{feydy19} showed the differentiability of $\ote^{\otimes}$ and the convexity of $\sdiv$ on measures with compact supports. On $\bbR^d$, more assumptions on $\alpha$ and $\beta$ are required. Throughout this section we restrict $\ote^{\otimes}$ and $\sdiv$ to the convex set of sub-Gaussian probability measures:
\begin{assumption}
	\label{ass:inputs}
	We set $C(x, y) = \|x - y\|^2$ and restrict $\ote^{\otimes}$ and $\sdiv$ to the set of sub-Gaussian probability measures
	$\cG(\bbR^d) \eqdef \{\mu | \exists q > 0,\, \bbE_\mu(e^{\frac{\|X\|^2}{2dq^2}}) \leq 2\}$.
\end{assumption}
\citet{mena19} showed that if $\alpha, \beta \in \cG(\bbR^d)$, there exists a pair of potentials $(f, g)$ verifying the fixed point equations \eqref{eq:schrodinger} on the whole space $\bbR^d$ that are bounded by quadratic functions. This result is key to show the differentiability of $\ote^{\otimes}$ on $\cG(\bbR^d)$. 
\begin{prop}[\citet{mena19}, Prop. 6]
	\label{prop:mena19}
	Let $\alpha, \beta \in \cG(\bbR^d)$. There exists a pair of smooth functions $(f, g)$ such that  \eqref{eq:schrodinger} holds on $\bbR^d$ and $\forall x, y \in \bbR^d$:
	\begin{align}
	\begin{split}
	\label{eq:bounds}
	- d q^2 (1 + \frac{1}{2}(\|x\| + \sqrt{2d}q)^2) \leq \frac{f(x)}{\varepsilon} \leq \frac{1}{2}(\|x\| + \sqrt{2d}q)^2 \\
		- d q^2 (1 + \frac{1}{2}(\|y\| + \sqrt{2d}q)^2) \leq \frac{g(y)}{\varepsilon} \leq \frac{1}{2}(\|y\| + \sqrt{2d}q)^2
		\end{split}
		\end{align}
\end{prop}
\paragraph{Differentiability}
In the rest of this section, $(f, g)$ denotes a pair of potentials defined by Proposition \ref{prop:mena19}.
We say that a function $F: \cG(\bbR^d) \to \bbR$ is differentiable at $\alpha$ if there exists $\nabla F(\alpha) \in \cC(\bbR^d)$ such that for any displacement $t\delta\alpha$ with $t>0$ and $\delta\alpha= \alpha_1 - \alpha_2$ with $\alpha_1, \alpha_2 \in \cG(\bbR^d)$,  and:
\begin{equation}
\label{eq:differentiability}
F(\alpha + t \delta\alpha)  = F(\alpha) + t\langle \delta\alpha, \nabla F(\alpha)\rangle + o(t)\enspace,
\end{equation}
where $\langle \delta \alpha,\nabla F(\alpha)\rangle = \int_{\bbR^d} \nabla F(\alpha) \diff \delta \alpha$.

\begin{prop}
\label{prop:gradient-ote}
Let $\alpha, \beta \in \cG(\bbR^d)$, and $(f, g)$ their associated pair of dual potentials given by proposition \ref{prop:mena19}. $\ote^{\otimes}(\alpha, .)$ is differentiable on sub-Gaussian measures with unbounded supports and its gradient is given by:
\begin{equation}
\label{eq:gradient-ote}
\nabla_{\beta}\ote^{\otimes}(\alpha, \beta) = g \enspace .
\end{equation}
\end{prop}
\proofsketch The proof is inspired from \citet{feydy19} in the case of measures with compact supports. The difference arises when taking the limit of integrals of the potentials. Thanks to assumption 1, proposition \ref{prop:mena19} provides an upper bound that allows to conclude by dominated convergence. The full proof is provided in the appendix.

The differentiability of $\sdiv$ follows immediately:
\begin{corollary}
	\label{cor:gradient-sdiv}
	\label{prop:gradient-sdiv}
Let $\alpha, \beta \in \cG(\bbR^d)$, and $(f, g)$ their associated pair of dual potentials given by proposition \ref{prop:mena19} and $h_\beta$ the autocorrelation potential associated with $\beta$. $\sdiv^{\otimes}(\alpha, .)$ is differentiable on sub-Gaussian measures with unbounded supports and its gradient is given by:
	\begin{equation}
	\label{eq:gradient-sdiv}
	\nabla_\beta\sdiv^{\otimes}(\alpha, \beta) =  g - h_\beta \enspace.
	\end{equation}
\end{corollary}
\begin{remark}
	It is important to keep in mind that the notion of differentiability (and gradient) of the functions $\ote^{\otimes}$ and $\sdiv$ differ from the usual Fr\'echet differentiability. Indeed, the space of probability measures $\cP(\bbR^d)$ has an empty interior in the space of signed Radon measures $\cM(\bbR^d)$. The definition adopted here defines derivatives along feasible directions in $\cP(\bbR^d)$. This is however sufficient to characterize the convexity of $\sdiv$ and its stationary points (see appendix \ref{ss:convexity} for details).
\end{remark}
\paragraph{Convexity}
Now we turn to showing that $\sdiv$ is convex with respect to either one of its arguments separately. To do so, we prove the first order characterization of convexity of a differentiable function $F: \cP_2(\bbR^d) \to \bbR$ given by:
\begin{equation}
\label{eq:convexity}
F(\alpha)  \geq F(\alpha') + \langle \alpha - \alpha', \nabla F(\alpha')\rangle \enspace,
\end{equation}
As shown by the proof of the following Lemma, the positivity of $K$ plays a key role in proving the convexity of $\sdiv$.
\begin{lemma}
	\label{lem:convexity}
	Let $\alpha, \alpha' \in \cG(\bbR^d)$ and let $h_\alpha, h_{\alpha'}$ denote their respective autocorrelation potentials given by proposition \ref{prop:symmetry}. Then if $\kernel(x, y) = e^{-\frac{\|x - y\|^2}{\varepsilon}}$:
	\begin{equation}
	\label{eq:bound-autocorr}
	\int e^{\frac{h_\alpha(x)}{\varepsilon}} \kernel(x, y)  e^{\frac{h_{\alpha'}(y)}{\varepsilon}} \diff \alpha(x) \diff\alpha'(y) \leq 1
	\end{equation}
\end{lemma}
\begin{prop}
	\label{prop:convexity}
Under assumption (1), $\sdiv$ is convex on sub-Gaussian measures with respect to either of its arguments.
\end{prop}
\proof
Let $\beta \in \cG(\bbR^d)$ . Let $\alpha, \alpha' \in \cG(\bbR^d)$. Let $(f, g)$ and $(f', g')$ denote the pair of potentials associated with $\ote^{\otimes}(\alpha, \beta)$ and $\ote^{\otimes}(\alpha', \beta)$ respectively and for any $\mu \in \cG(\bbR^d)$, let  $h_\mu$ denote the autocorrelation potential associated with $\ote^{\otimes}(\mu, \mu)$. The first order inequality \eqref{eq:convexity} applied to $F = S_{\varepsilon}(., \beta)$ is equivalent to:
\begin{align}
\label{eq:align-convex}
\begin{split}
&\eqref{eq:convexity} \Leftrightarrow\langle \alpha, f - h_\alpha\rangle + \langle \beta, g - h_{\beta} \rangle \geq \\ &\langle  \alpha', f' - h_{\alpha'}\rangle + \langle \beta, g' - h_{\beta} \rangle 
+ \langle \alpha - \alpha', f' - h_{\alpha'}\rangle \\
&\Leftrightarrow \langle \alpha, f - h_\alpha\rangle + \langle \beta, g \rangle \geq \langle \beta, g'  \rangle 
+ \langle \alpha , f' - h_{\alpha'}\rangle \\
&\Leftrightarrow \langle \alpha, f\rangle + \langle \beta, g \rangle \geq \langle \beta, g'  \rangle 
+ \langle \alpha , f' - h_{\alpha'} +  h_\alpha\rangle \\
&\Leftrightarrow \ote^{\otimes}(\alpha, \beta) \geq \langle \alpha , f' - h_{\alpha'} +  h_\alpha\rangle + \langle \beta, g'  \rangle 
\end{split}
\end{align}
To show the last inequality we use the definition of the dual problem \eqref{eq:otdual} and evaluate the dual function at the suboptimal potentials $(f' - h_{\alpha'} +  h_\alpha,  g')$. Doing so leads to:
\begin{align*}
&\ote^{\otimes}(\alpha, \beta) \geq \langle \alpha , f' - h_{\alpha'} +  h_\alpha\rangle + \langle \beta, g'  \rangle +  \varepsilon  \\& - \varepsilon \int_{\bbR^{d} \times \bbR^{d}} \exp\left(\frac{(f' - h_{\alpha'} +  h_\alpha) \oplus g' - C}{\varepsilon}\right) \diff \alpha \diff \beta \enspace .
\end{align*}
To conclude, all we need to show is that,
\begin{equation}
 \int_{\bbR^{d} \times \bbR^{d}} \exp\left(\frac{(f' - h_{\alpha'} +  h_\alpha)  \oplus g' - C}{\varepsilon}\right) \diff\alpha\diff\beta\leq 1
\end{equation}
By the Fubini-Tonelli theorem, the order of integration is irrelevant. First integrating with respect to $\beta$, we use the optimality conditions \eqref{eq:schrodinger} on the pair $(f', g')$ then on $h_{\alpha'}$:
\begin{align*}
B =& \int_{\bbR^{d} \times \bbR^{d}} \exp\left(\frac{(f' - h_{\alpha'} +  h_\alpha) \oplus g' - C}{\varepsilon}\right) \diff \alpha \diff \beta \\
=& \int_{\bbR^{d}} \exp\left(\frac{h_\alpha- h_{\alpha'}  }{\varepsilon}\right) \diff \alpha \\
=& \int_{\bbR^{d} \times \bbR^{d}} \exp\left(\frac{h_\alpha \oplus h_{\alpha'}  - C}{\varepsilon}\right) \diff \alpha \diff \alpha'
\end{align*}
Thus, Lemma \ref{lem:convexity} applies and we have $ B \leq 1$. \qed
\paragraph{Barycenter of sub-Gaussian distributions.}
 We have shown that $\ote^{\otimes}$ and $\sdiv$ are convex and differentiable, thus the weighted barycenters $\alpha_{\ote^\otimes}$ and $\alpha_{\sdiv}$ can be characterized by the first order optimality condition as follows. Let $(f_k, g_k)$ denote the potentials associated with $\ote^{\otimes}(\alpha_k, \alpha)$ and $h_{\alpha}$ the autocorrelation potential associated with $\ote^{\otimes}(\alpha, \alpha)$. Using the first order characterization of convexity \eqref{eq:convexity}, $\alpha^\star$ is a global minimizer of the barycenter loss of $\ote^\otimes$ if and only if for any direction $\beta \in \cG(\bbR^d)$, $\langle \sum_{k=1}^K w_k {\nabla_{\alpha^\star} \ote^{\otimes}}(\alpha_k, \alpha^\star), \beta - \alpha^\star\rangle \geq 0$. This is equivalent to $\sum_{k=1}^K w_k\langle g_k, \beta - \alpha^\star\rangle\geq 0\enspace$. Similarly, for $\alpha_{\sdiv}$ we get the optimality condition
$\sum_{k=1}^K w_k\langle g_k - h_{\alpha^\star}, \beta - \alpha^\star\rangle\geq 0\enspace$.
We are now ready to summarize the different steps of the proofs of the theorems. For $\sdiv$, we provide solutions of the optimality conditions by considering quadratic potentials and Gaussian barycenters $\alpha_{\sdiv}$. We proceed by identification of the coefficients of the polynomials and the parameters of the barycenters and show that the obtained solutions verify the optimality condition. For $\ote^{\otimes}$, we proceed similarly for $2\varepsilon'^2 < \sigma^2$. For $2\varepsilon'^2 \geq 2\sigma^2$, we show directly that for the Dirac measure $\alpha^\star = \delta_{\bar{\mu}}$, there exists a set of potentials that verify the optimality condition alongside Sinkhorn's equations. The detailed derivations are provided in the supplementary materials.
\section{Fast Sinkhorn-like algorithm}
\label{s:algorithm}
\begin{figure}[t]
	\centering
	\vspace{-0.2cm}
	\includegraphics[width=0.8\linewidth, trim={1.cm 0 0.cm 0.cm}, clip]{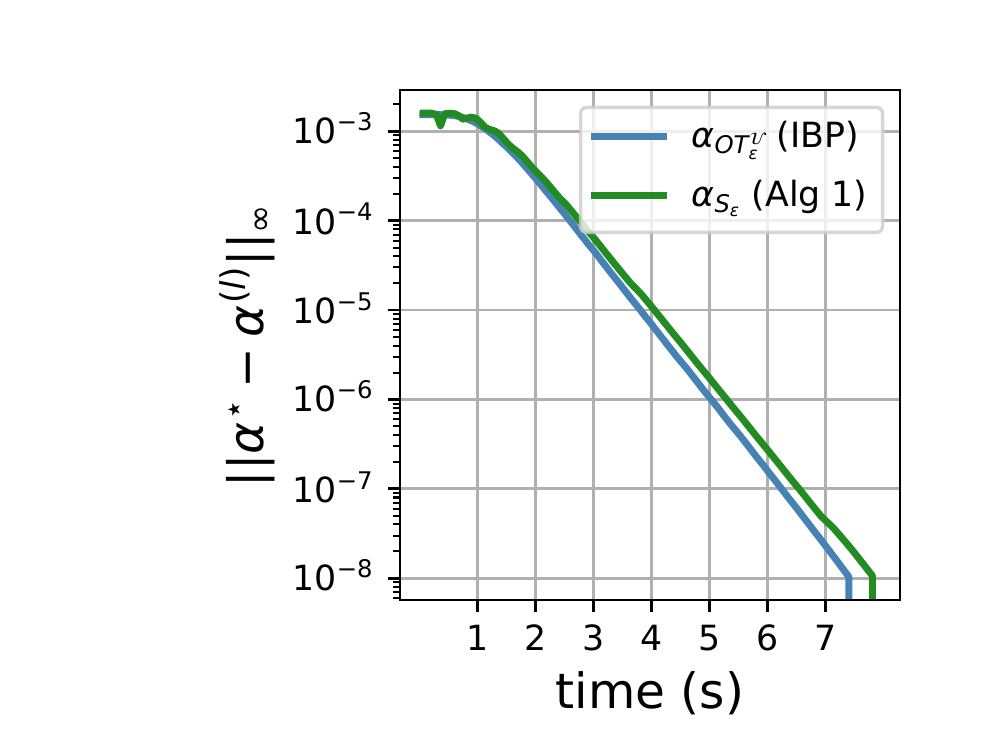}
	\vspace{-0.4cm}
	\caption{Convergence to the true barycenters of univariate Gaussians $\cN(-0.5, 0.1)$ and $\cN(0.5, 0.1)$. Algorithm \ref{alg:debiased-sinkhorn} is as fast as IBP with a linear convergence rate. \label{f:convergence}}
\end{figure}

\paragraph{Discrete measures on a finite space}
The purpose of this section is to derive a fast Sinkhorn-like algorithm to compute $\alpha_{\sdiv}$ on a fixed support. Let $\cX = \{x_1, \dots, x_n\}$ be a finite grid of size $n$. With images for instance, each $x_i$ would correspond to a pixel.  We identify a probability measure $\alpha = \sum_{i=1}^n  \alpha_i \delta_{x_i} \in \cP(\cX)$ with its weights vector $(\alpha_i) \in \bbR^n_{++}$ such that $\sum_{i=1} \alpha_i = 1$. In the rest of this paper, $\ote$ and $\sdiv$ can be seen as functions operating on the interior of the probability simplex of $\bbR^n$ denoted by $\Delta_n = \{x \in \bbR^n_{++} | \sum_{i=1}x_i = 1\}$. We assume that the cost matrix $\bC \in \bbR^{n\times n}_+$ is symmetric negative semi-definite (or equivalently, its associated kernel $\bK = e^{-\frac{\bC}{\varepsilon}}$ is positive semi-definite). This assumption holds for instance if $ \bC_{ij} = \|x_i - x_j\|^p$ with $p \in ]0, 2]$ (see ~\citep[§3, Thm 2.2, Cor 3.3]{berg84} for both claims)
\paragraph{Debiased barycenters} To obtain a fast iterative algorithm for the debiased barycenters $\alpha_{\sdiv}$, we are going to leverage the IBP algorithm through the uniform measure on $\cX$ as follows. First, the identity \eqref{eq:sinkhorn-div-ref} ensures that $\sdiv$ is independent of the reference measures. Thus, one can write:
\begin{equation*}
\label{eq:sinkhorn-div-unif-alg}
\sdiv(\alpha, \beta) =  \ote^\cU(\alpha, \beta) - \frac{\ote^\cU(\alpha, \alpha) + \ote^\cU(\beta, \beta)}{2}\enspace.
\end{equation*}
Using \eqref{eq:otdef-uniform-kl}, one can write $\ote^\cU(\alpha, \beta) $ as  a $\kl$ projection. The remaining autocorrelation terms can be replaced by their dual problems to obtain the following proposition. A detailed derivation is provided in appendix \ref{ss:proof-alg1}.
\begin{prop}
	\label{prop:sdiv-kl} Let $\alpha_1, \dots, \alpha_K \in \Delta_n$ and $\bK = e^{-\frac{\bC}{\varepsilon}}$. Let $\pi$ denote a sequence $\pi_1, \dots, \pi_K$ of transport plans in $\bbR_+^{n\times n}$ and the constraint sets $\cH_1=\{ \pi | \forall k, \, \pi_k \mathds 1 = \alpha_k\}$, and $\cH_2 = \{ \pi | \forall k\, \forall k', \, \pi_k^\top \mathds 1 = \pi_{k'} \mathds 1 \}$.
The barycenter problem $\min_{\alpha \in \Delta_n} \sum_{k=1}^K w_k \sdiv(\alpha_k, \alpha)$ is equivalent to:
\begin{align}
\label{eq:sdiv-bar-kl}
\begin{split}
 \min_{\substack{\pi \in \cH_1 \cap \cH_2 \\  d \in \bbR_+^n}} & \Bigg [ \varepsilon \sum_{k=1}^K w_k \widetilde{\kl}(\pi_k| \bK \diag(d)) \\
& + \frac{\varepsilon}{2} \langle d - \mathds 1 , \bK (d - \mathds 1) \rangle \Bigg ] \enspace .
\end{split}
\end{align}
where $\widetilde{\kl}(\bA, \bB) = \sum_{i,j}^n\bA_{ij} \log\left(\frac{\bA_{ij}}{\bB_{ij}}\right) + \bB_{ij} - \bA_{ij}$.
\end{prop}
Since $\widetilde{\kl}$ is jointly convex and $\bK$ is assumed positive-definite, the objective \eqref{eq:sdiv-bar-kl} is convex. Minimizing \eqref{eq:sdiv-bar-kl} with respect to $\pi$ leads to the barycenter problem $\alpha_{\ote^{\cU}}$ \eqref{eq:ibp} with the modified kernel $\bK \diag(d)$. This problem can be solved via the fast IBP algorithm. 
Minimizing with respect to $d$ leads to the Sinkhorn fixed point equation $d = \frac{\sum_{w_k} \pi_k^\top \mathds 1}{\bK d}$ for which there exists a converging sequence $d_{n+1} \gets \sqrt{\frac{ d_n \odot \sum_{w_k} \pi_k^\top \mathds 1}{\bK d}} (\star) $~\citep{knight14}. Given that \eqref{eq:sdiv-bar-kl} is smooth and convex, alternate minimization -- which amounts to perform IBP and (*) iterations -- converges towards its minimum.
However, we notice that in practice, either taking one iteration or fully optimizing the subproblems produces the same minimizer. We thus propose to combine one IBP iteration with the update ($\star$), which leads to Algorithm \ref{alg:debiased-sinkhorn} (see the appendix for further details on the IBP algorithm).  Using the theoretical barycenters of Gaussians given by theorems \ref{thm:lebesgue} and \ref{thm:div}, we can monitor the convergence to the ground truth (Figure \ref{f:convergence}). Theoretically, both IBP and algorithm \ref{alg:debiased-sinkhorn} have a $\cO(Kn^2)$ complexity per iteration. A convergence proof of IBP can obtained using alternating Bregman projections (See \cite{benamou14} and the references theirein). For Algorithm \ref{alg:debiased-sinkhorn} however, similar techniques were not successful. Proving its convergence will be pursued in future work.
\begin{algorithm}[t]
	\caption{Debiased Sinkhorn Barycenter}
	\label{alg:debiased-sinkhorn}
	\begin{algorithmic}
		\STATE {\bfseries Input:}  $ \alpha_1, \dots, \alpha_K$,  $\bK = e^{-\frac{\bC}{\varepsilon}}$
		\STATE {\bfseries Output:} $\alpha_{\sdiv}$ 
		\STATE Initialize all scalings $(b_k), d$ to $ \mathds 1$, 
		\REPEAT
		\FOR{$k=1$ {\bfseries to} $K$}
		\STATE $a_k \gets \left(\frac{\alpha_k}{\bK b_k}\right)$
		\ENDFOR
		\STATE $\alpha \gets d \odot \prod_{k=1}^K ( \bK^\top a_k) ^{ w_k } $
		\FOR{$k=1$ {\bfseries to} $K$}
		\STATE $b_k \gets \left(\frac{\alpha}{\bK^\top a_k}\right)$
		\ENDFOR
		\STATE $d \gets \sqrt{d\odot\left( \frac{\alpha }{\bK d}\right)}$
		\UNTIL{convergence}
	\end{algorithmic}
\end{algorithm}
\section{Applications}
\label{s:experiments}
\begin{figure}[t!]
	\includegraphics[width=\linewidth]{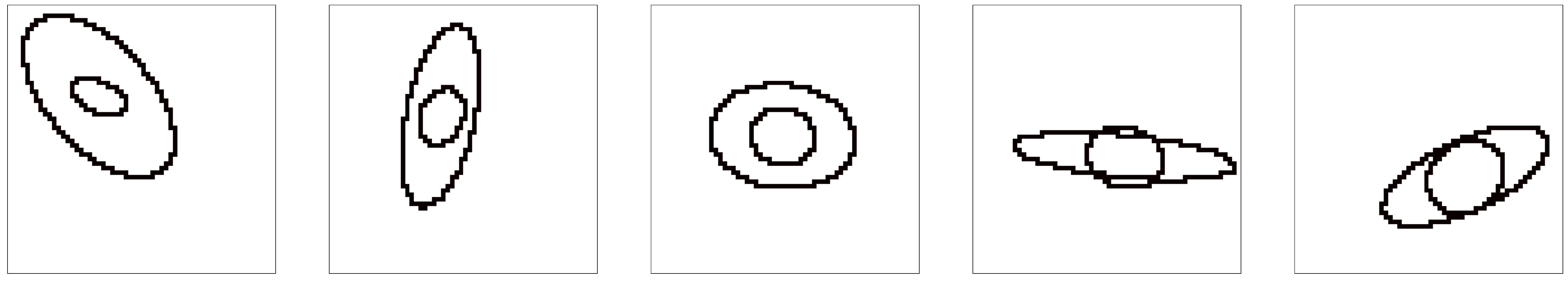}
	\vspace{-0.6cm}
	\caption{5 examples of random nested ellipses of size (60 $\times$ 60) used to compute the barycenters of Figure \ref{f:bar-ellipses}.  \label{f:ellipses}}
\end{figure}

Now we turn to showing the practical benefits of debiased barycenters in terms of accuracy, speed and performance.
\paragraph{Benchmarks}
In addition to $\alpha_{\ote^\cU}$, $\alpha_{\ote^{\otimes}}$, we evaluate the performance of the following barycenters: 
\begin{itemize}
\item $\alpha_{A_\varepsilon}$: Sharp barycenters introduced by \citet{luise18}, where $A_\varepsilon$ is defined as:
$A_\varepsilon(\alpha, \beta) = \langle \bC, \pi_\varepsilon^\star(\alpha, \beta)\rangle$. Here $\pi_\varepsilon^\star(\alpha, \beta)$ is the primal minimizer of the regularized problem $\ote^\cU(\alpha, \beta)$, computed via accelerated gradient descent.
\item $\alpha_{\sdiv}^{F}$: Free support barycenters introduced by \citet{luise19} that uses the same debiased divergence $\sdiv$, and deals with the free support problem by adding / removing a Dirac particle with Frank-Wolf's algorithm.
\item $\alpha_{W}$: The original non-regularized Wasserstein problem solved with interior point methods - using the accelerated MAAIPM algorithm of \citet{dongdong19}.
\end{itemize}
\paragraph{Debiased barycenters of ellipses}
To demonstrate how debiased barycenters $\alpha_{\sdiv}$ reduce smoothing and are computationally competitive with $\alpha_{\ote^\cU}$, we compare the barycenters of 10 randomly generated nested ellipses displayed in Figure \ref{f:ellipses}. We set the cost matrix $\bC$ to the squared Euclidean distance on the unit square and set $\varepsilon = 0.002$. We use the same termination criterion for all methods based on a maximum relative change of the barycenters set to $10^{-5}$.

For $\alpha_{\sdiv}, \alpha_{\ote^\cU}, \alpha_{\ote^{\otimes}}, \alpha_{A_\varepsilon}$, we use the convolution trick introduced by \citet{solomon15} which amounts to computing the kernel operation $\bK a$ on a vectorized image $a$ by applying a Gaussian convolution on the rows and the columns of $a$, thereby reducing the complexity of one Debiased / IBP iteration from $O(n^2)$ to $O(n^{\frac{3}{2}})$. 

\begin{figure}[t!]	
	\includegraphics[width=\linewidth]{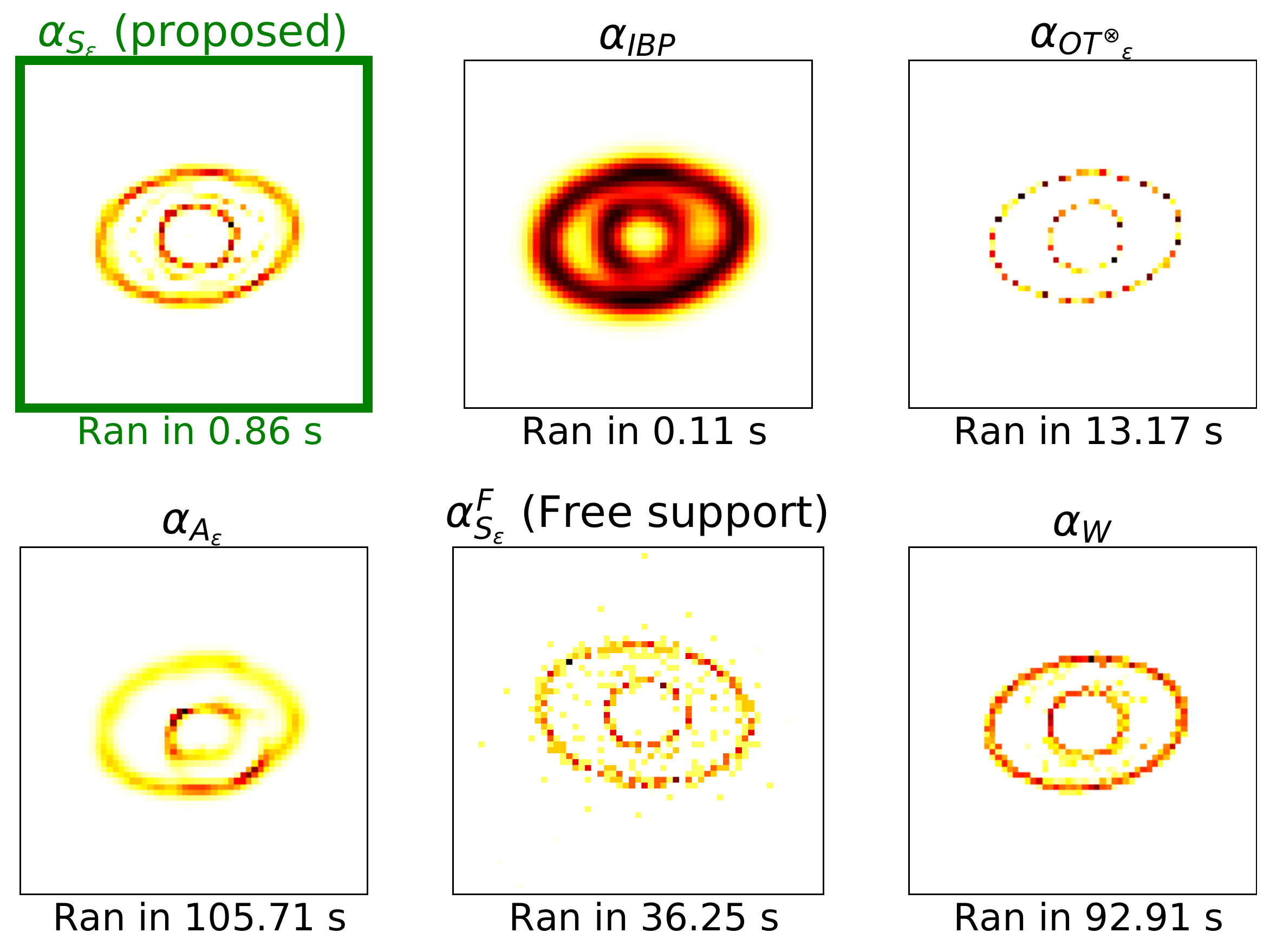}
	\caption{Barycenters of the 10 nested ellipses shown in Figure \ref{f:ellipses}.
	Results illustrate the reduced blurring of the proposed approach and 
	running times presented below each image demonstrate the computational
	efficiency.
	All 6 barycenters were computed on a laptop with an Intel Core i5 3.1 GHz Processor.\label{f:bar-ellipses}}
\end{figure}
Figure \ref{f:ellipses} shows that even though $\alpha_{A_\varepsilon}$ and $\alpha_W$ are not blurred
compared to $\alpha_{\ote^{\cU}}$, they cannot compete computationally with Sinkhorn-like algorithms. The debiased barycenter is sharp and runs in about the same time as $\alpha_{\ote^{\cU}}$.
Besides, the shrinking bias of $\ote^{\otimes}$ unfolded by theorem \ref{thm:product} is illustrated in the degeneracy of the ellipse $\alpha_{\ote^{\otimes}}$.

\begin{figure}[!t]
\includegraphics[width=0.19\linewidth, trim={7cm 0 7cm 0}, clip]{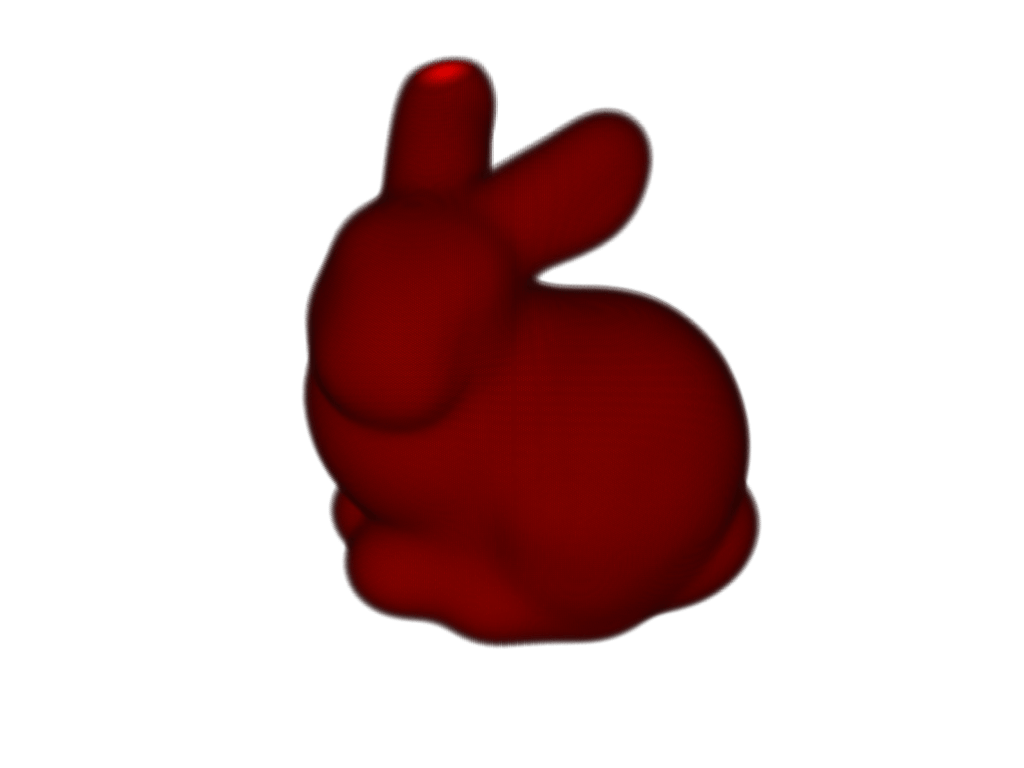}
\includegraphics[width=0.19\linewidth, trim={7cm 0 7cm 0}, clip]{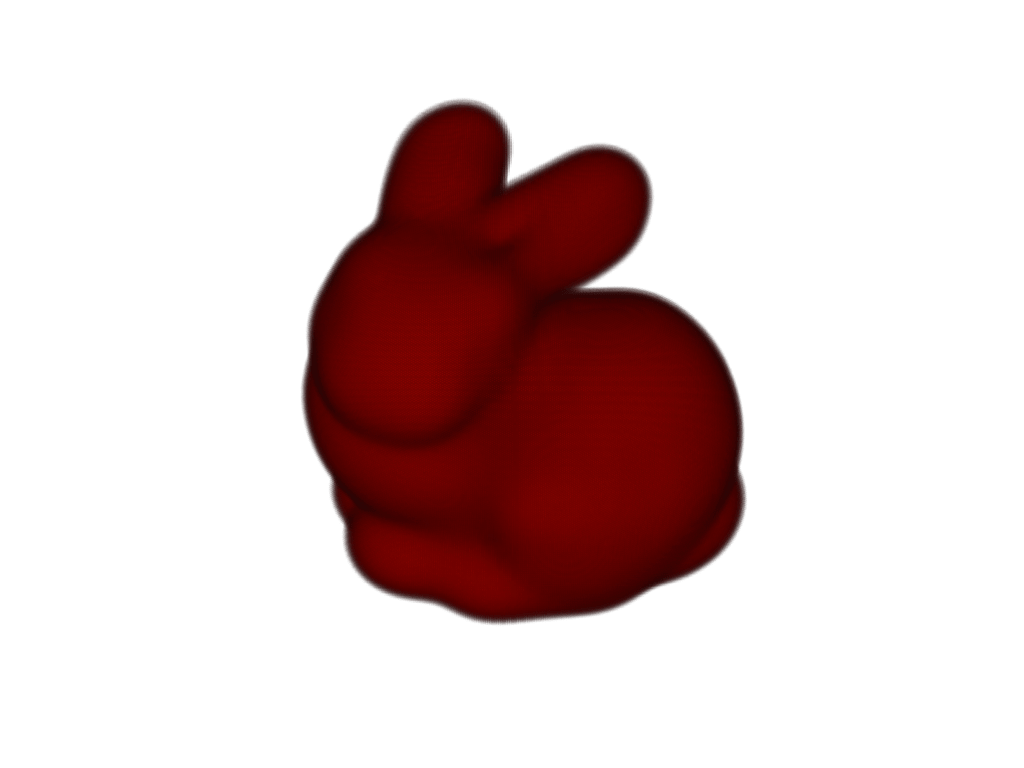}
\includegraphics[width=0.19\linewidth, trim={7cm 0 7cm 0}, clip]{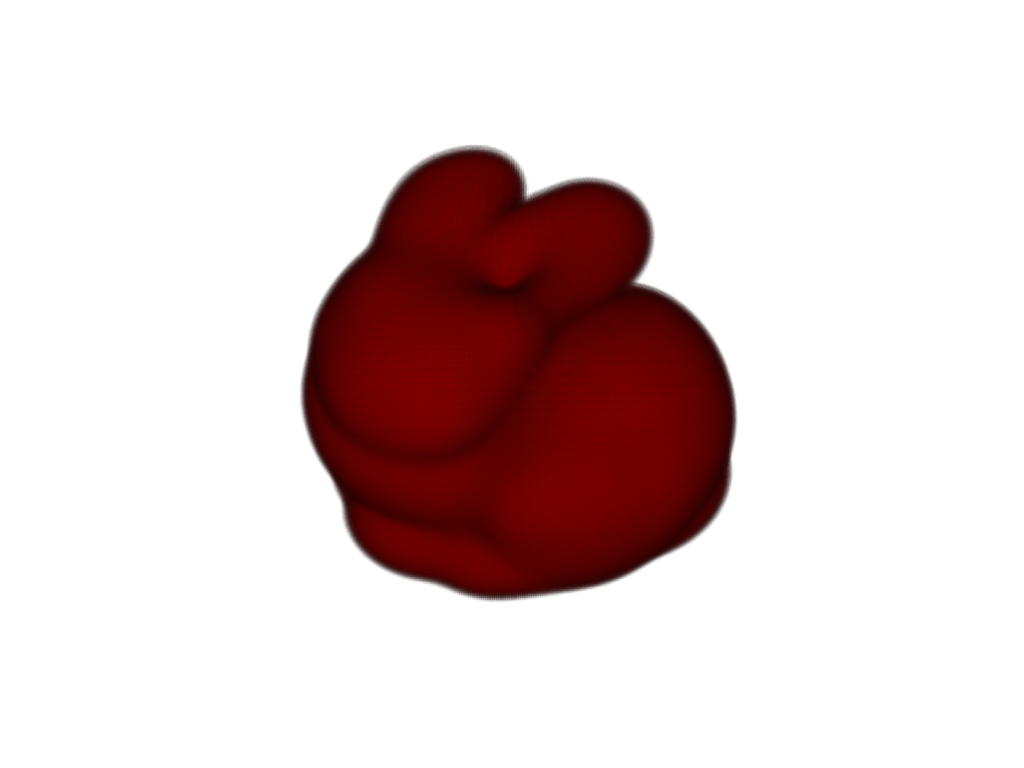}
\includegraphics[width=0.19\linewidth, trim={7cm 0 7cm 0}, clip]{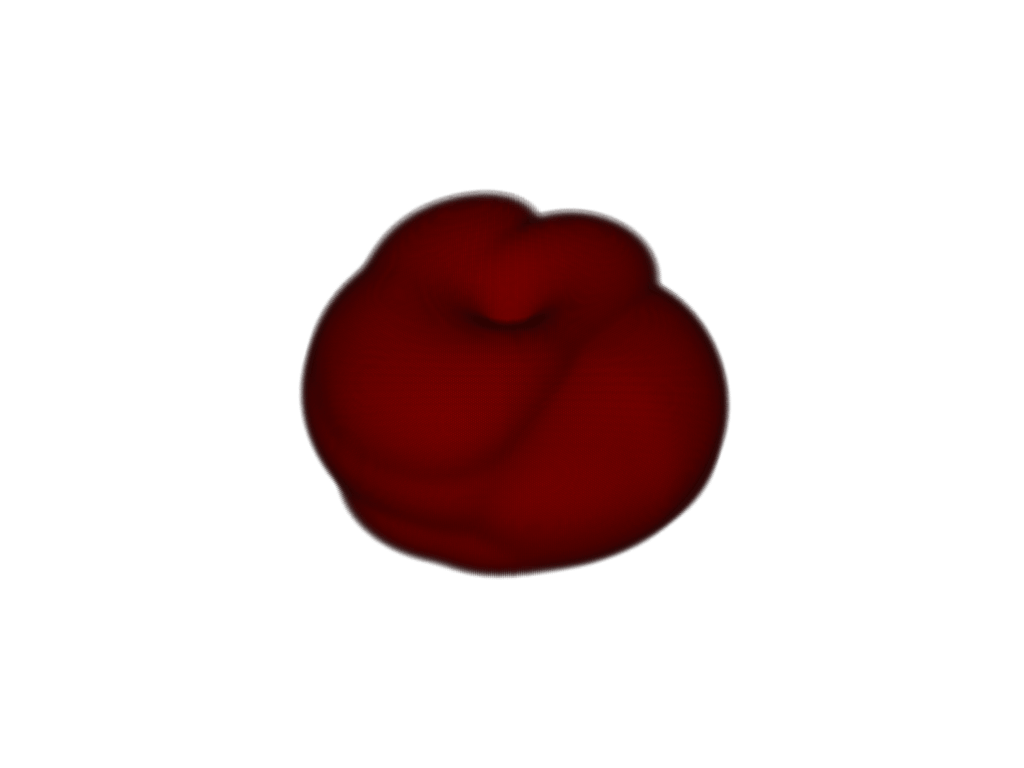}
\includegraphics[width=0.19\linewidth, trim={7cm 0 7cm 0}, clip]{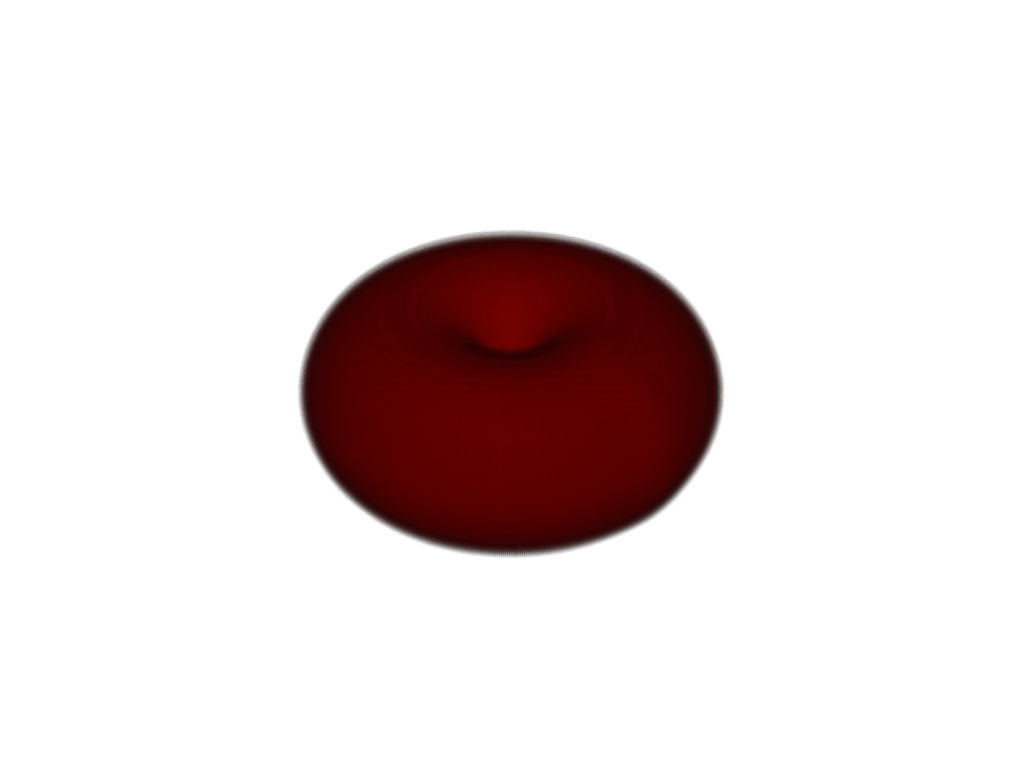}

\vspace{-0.6cm}
\caption{Interpolation of two 3D shapes a $(200)^3$ uniform grid with IBP illustrating a clear blurring bias of $\ote^{\cU}$. \label{f:rabbit-ibp}}
\end{figure}

\begin{figure}[!t]
	\includegraphics[width=0.19\linewidth, trim={7cm 0 7cm 0}, clip]{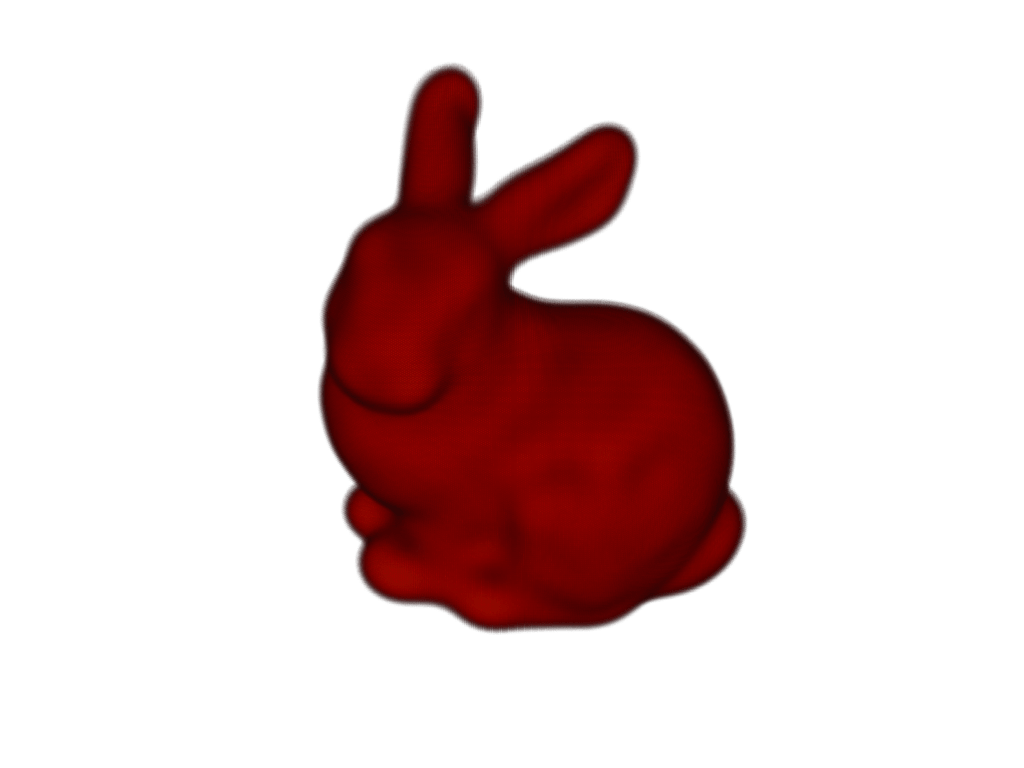}
	\includegraphics[width=0.19\linewidth, trim={7cm 0 7cm 0}, clip]{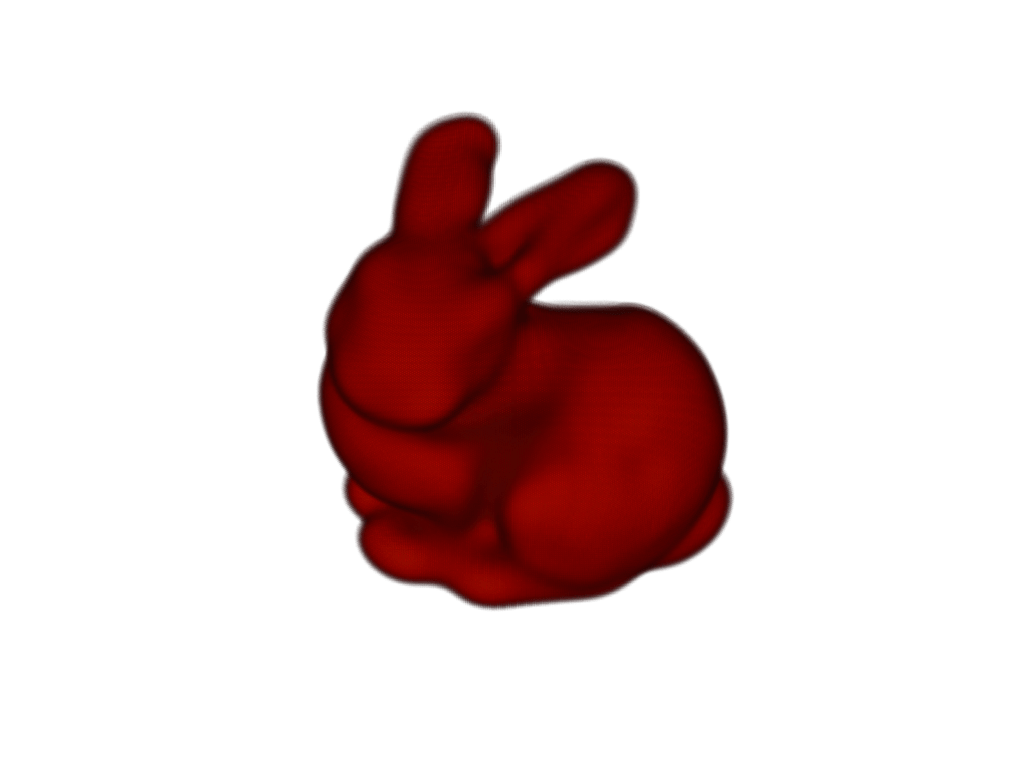}
	\includegraphics[width=0.19\linewidth, trim={7cm 0 7cm 0}, clip]{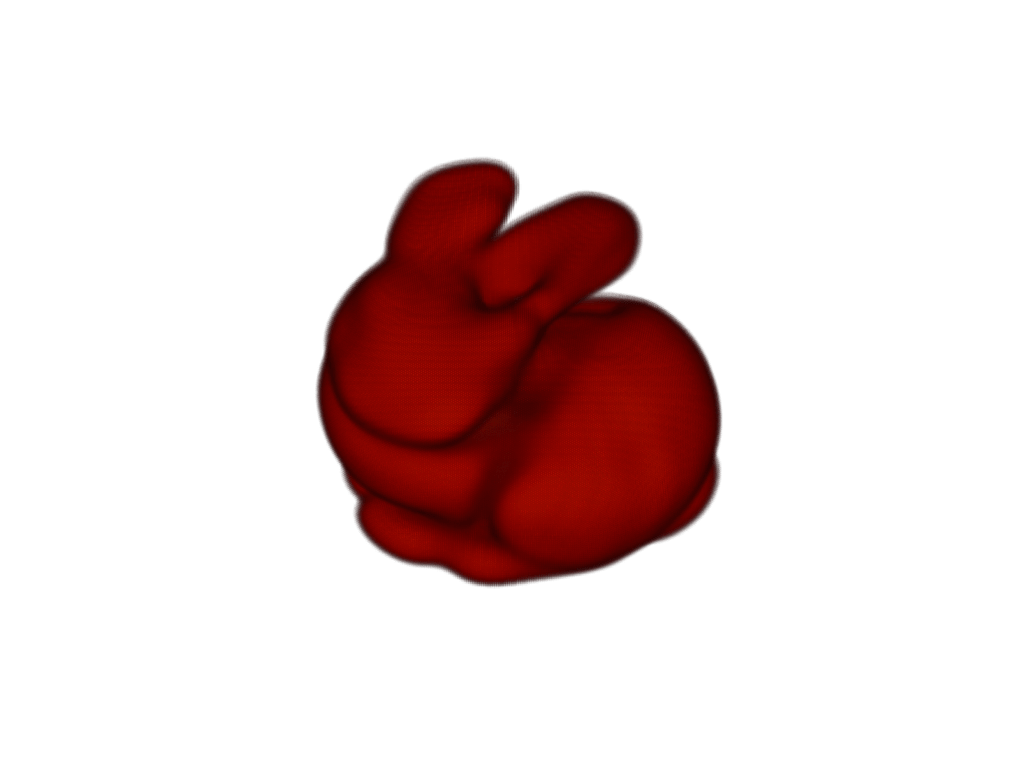}
	\includegraphics[width=0.19\linewidth, trim={7cm 0 7cm 0}, clip]{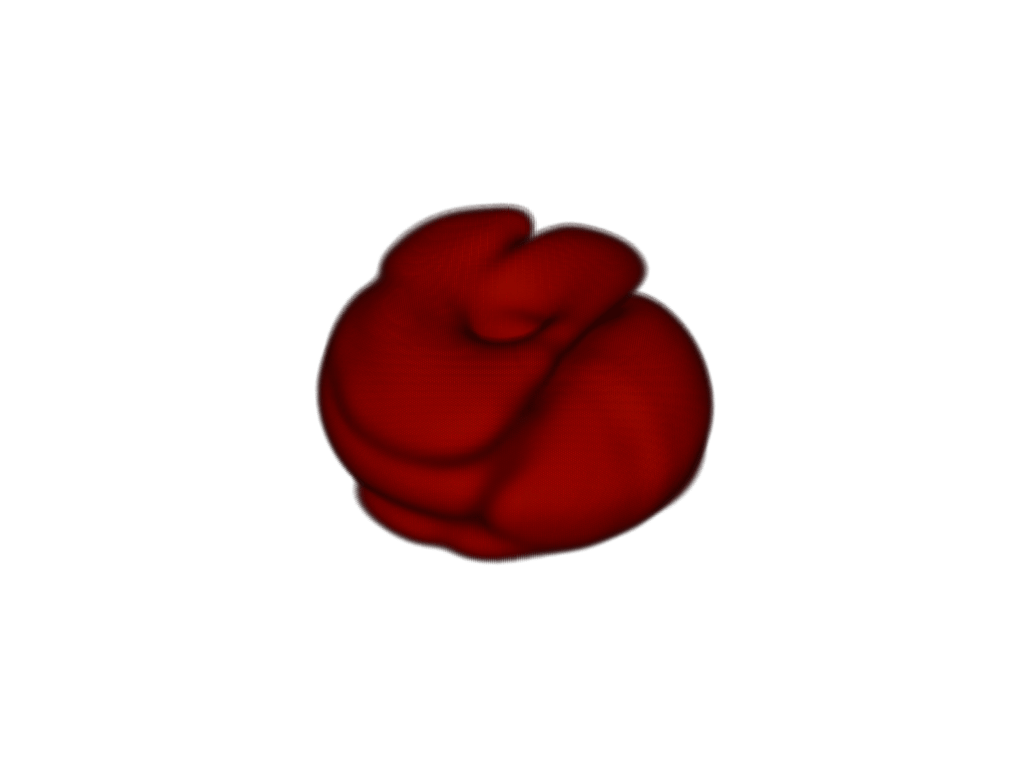}
	\includegraphics[width=0.19\linewidth, trim={7cm 0 7cm 0}, clip]{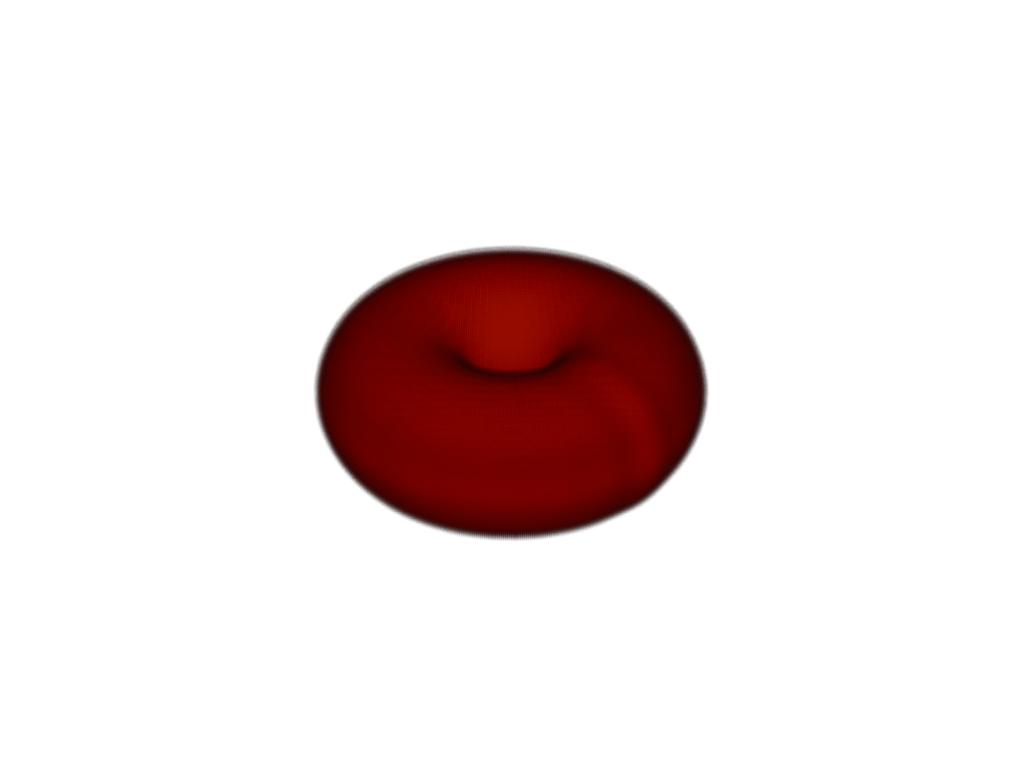}
	\vspace{-0.6cm}
	\caption{Interpolation of two 3D shapes on a $(200)^3$ uniform grid with the proposed Debiased Sinkhorn (Alg \ref{alg:debiased-sinkhorn}). The interpolation is sharper and completes in about the same time as figure \ref{f:rabbit-ibp} (5 seconds on a GPU).  \label{f:rabbit-deb}}
\end{figure}

\paragraph{Barycenters of 3D shapes} To visually illustrate the impact of the reduced smoothing bias of $\sdiv$,
we computed a barycentric interpolation of shapes discretized
in a 3D grid of $200\times 200\times200$ voxels. The different interpolations correspond to weights $(w, 1-w)$ where $w \in [0, 0.25, 0.5, 0.75, 1]$.  We set the cost matric $\bC$ to the squared Euclidean distance on the unit cube and set $\varepsilon = 0.01$. Results presented in Figures \ref{f:rabbit-ibp} and \ref{f:rabbit-deb} using $\ote^{\cU}$
and $\sdiv$ qualitatively demonstrate that $\sdiv$ leads to sharper
edges, while in both cases it takes a few seconds to compute on a GPU.
Again, the kernel operation $\bK a$ on a vectorized 3D grid $a$ can be computed via a sequence of 3 Gaussian convolutions
on each axis $(x, y, z)$ which reduces the complexity of one Debiased / IBP iteration from
$O(n^2)$ to $O(n^{\frac{4}{3}})$.

\begin{figure}[!t]
	\centering
		\includegraphics[width=0.9\linewidth]{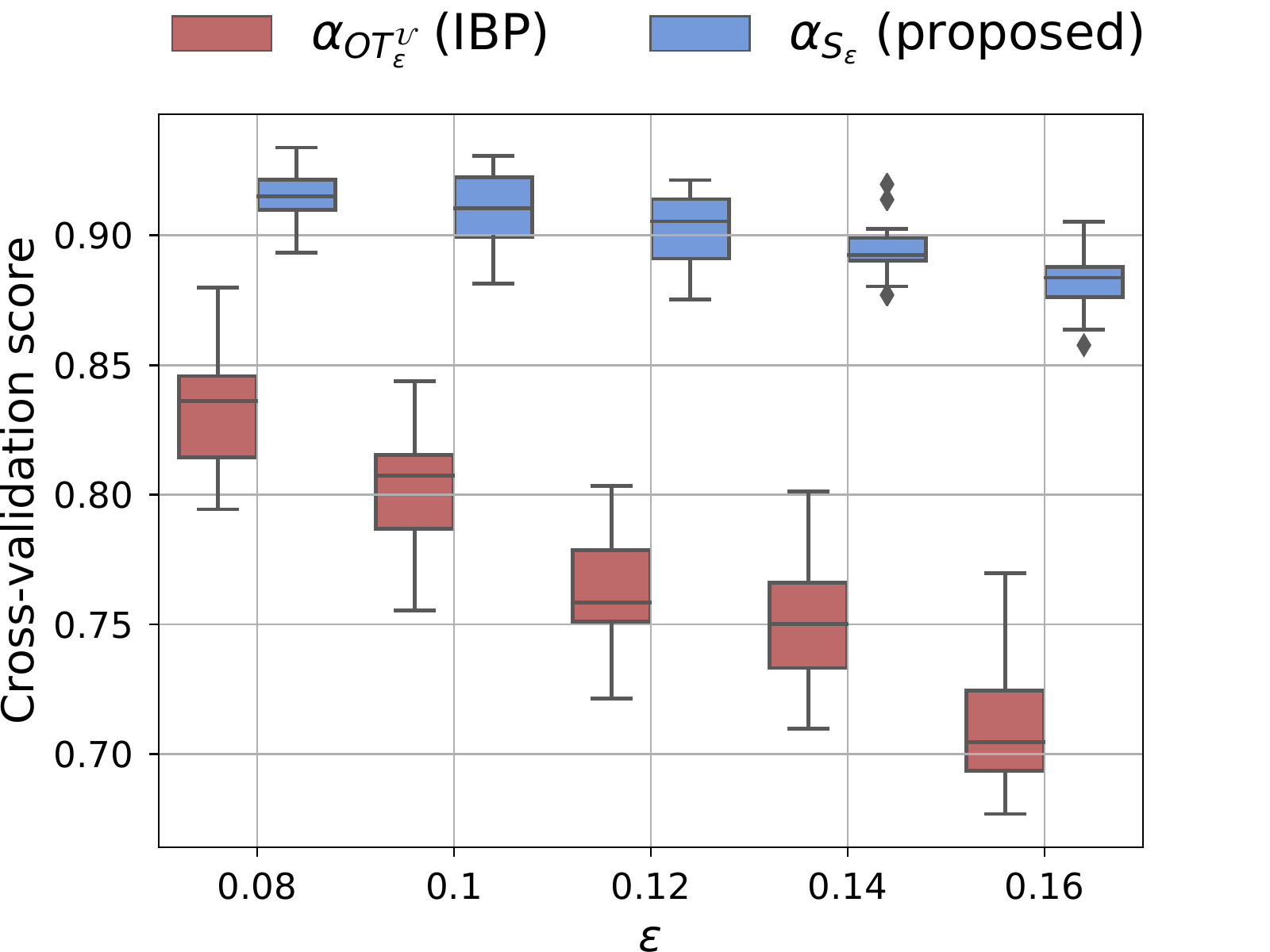}
		\caption{Cross-validation accuracy with 95\% confidence intervals obtained
		on 500 MNIST images using barycentric embedding with $\sdiv$
		or $\ote^{\cU}$. Debiasing of $\sdiv$ improves performance.
		$\sdiv$ is less sensitive to $\varepsilon$. \label{f:mnist-cv}}
\end{figure}

\paragraph{Optimal transport barycentric embeddings}
One of the many machine learning applications of OT barycenters is to compute low-dimensional barycentric embeddings. Introduced by \citet{bonneel16}, OT barycentric coordinates are defined as follows. Given a dictionary $\cA$ of distributions $\alpha_1, \dots, \alpha_K$ and $w \in \Delta_K$, let $\alpha_F(w) = \argmin_{\alpha}\sum_{k=1}^K w_kF(\alpha_k, \alpha)$ for some OT divergence $F$. The OT coordinates $\hat{w}$ of a distribution $\beta$ are defined as the weights of the barycenter $\alpha_F(w)$ best approximating $\beta$ for a given divergence.  Using a quadratic divergence, it reads: $\hat{w} = \argmin_{w\in \Delta_K} \|\alpha_F(w) - \beta\|^2$. To leverage the differentiability of the IBP iterations, \citet{bonneel16} used the divergence $\ote^{\cU}$ and proposed to substitute the minimizer $\alpha_F(w)$ with the $l$-th IBP iterate $\alpha_F^{(l)}(w)$. Differentiating the barycenter nets $\alpha_F^{(l)}(w)$ with respect to $w$ can be done via automatic differentiation, while the full minimization can be done using accelerated gradient descent using a soft-max reparametrization. Here we use the ADAM optimizer of the pyTorch library \citep{torch}. To evaluate the benefits of debiasing, we take 500 samples of the MNIST dataset \citep{mnist} with 100 instances of each digit (0-1-2-3-4). We select 10\% of the dataset (a subset of 50 images; ergo K=50) at random as our learning dictionary $\cA$ and compute the barycentric coordinates of the remaining 90\% subset denoted as $\cD$. Thus, for each image among the 450 samples of $\cD$,  we compute the closest (in squared $\ell_2$) weighted barycenter of the elements of $\cA$ by optimizing over the weights. Thus, each image is represented by a vector of weights $w \in \Delta_K$. Our new embedded dataset is now a table of shape $(450 \times 50)$. We train a random forest classifier using the Scikit-learn library \citep{sklearn} on this learned embedding) and compute a 10-fold cross-validation. Figure \ref{f:mnist-cv} displays the accuracy scores for $F = \ote^{\cU}$ and $F=\sdiv$ for 20 different randomized selections of the dictionary $\cA$. The debiased $\sdiv$ improves accuracy and is less sensitive to the setting of $\varepsilon$.

\section*{Conclusion}
Entropy regularized OT was previously known to induce a bias that can be mitigated using Sinkhorn divergences. Using OT barycenters of Gaussian distributions, we have shown that this entropy bias can be a blur or a shrink depending on the reference measure defining the relative entropy function. We have also extended the convexity and differentiability properties of OT and the Sinkhorn divergence to measures with non-compact supports.

\section*{Acknowledgments}
MC and HJ acknowledge the support of a chaire
d’excellence de l'IDEX Paris Saclay. AG and HJ were supported by the European Research Council Starting
Grant SLAB ERC-YStG-676943. We thank Thibault S\'ejourn\'e and Fran\c{c}ois-Xavier Vialard for fruitful discussions, in particular for pointing out the identity \eqref{eq:reference-l}. We thank Zikai Ziong for sharing the matlab code and adapting it to our ellipses experiment.

\bibliography{references}
\bibliographystyle{icml2020}

\iftoggle{supplementary}{
	\cleardoublepage

\appendix
\onecolumn

\section{Convexity and Optimality condition}
\label{ss:convexity}
In this section we show how the notion of differentiability along feasible directions in $\cP(\bbR^d)$ is enough to characterize convexity and first order optimality conditions. Consider an arbitrary function $F$ on the space of probability measures.

\begin{definition}
\label{sdef:differentiability}
 F is said to be differentiable at $\alpha\in \cP(\bbR^d)$, if and only if there exists $\nabla F(\alpha) \in \cC(\bbR^d)$ such that for any displacement $\delta\alpha= \alpha_1 - \alpha_2$ with $\alpha_1, \alpha_2 \in \cP(\bbR^d)$:
\begin{equation}
\label{seq:differentiability}
F(\alpha + t \delta\alpha)  = F(\alpha) + t\langle \delta\alpha, \nabla F(\alpha)\rangle + o(t)\enspace,
\end{equation}
where $\langle \eta,\nabla F(\alpha)\rangle = \int_{\bbR^d} \nabla F(\alpha) \diff \eta$.
\end{definition}

\begin{prop}[convexity]
	\label{sprop:convexity}
Assume F is differentiable on $\cP(\bbR^d)$. $F$ is convex If and only if for all $\alpha, \alpha' \in \cP(\bbR^d)$:
\begin{equation}
\label{seq:convexity}
F(\alpha)  \geq F(\alpha') + \langle \alpha - \alpha', \nabla F(\alpha')\rangle \enspace,
\end{equation}
\end{prop}
\proof
($\Rightarrow$). Assume \eqref{seq:convexity} holds.
Let $\lambda \in [0, 1]$ and $\alpha_\lambda = \lambda \alpha + (1-\lambda \alpha')$ with arbitrary probability measures $\alpha, \alpha'$. Applying \eqref{seq:convexity} twice with $\alpha' = \alpha_\lambda$ leads to:
\begin{align*}
F(\alpha)  \geq F(\alpha_\lambda) + \langle \alpha - \alpha_\lambda, \nabla F(\alpha_\lambda)\rangle \\
F(\alpha')  \geq F(\alpha_\lambda) + \langle \alpha' - \alpha_\lambda, \nabla F(\alpha_\lambda)\rangle
\end{align*}
Multiplying the first equation by $\lambda$ and the second one by $1 - \lambda$ before summing leads to:
$$ \lambda F(\alpha)  + (1-\lambda)F(\alpha') \geq F(\alpha_\lambda). $$
Thus F is convex.

($\Leftarrow$). Assume F is convex. Let $\lambda \in (0, 1)$. Convexity implies that:
\begin{align*}
&F(\lambda \alpha +(1-\lambda)\alpha' ) \leq \lambda F(\alpha)  + (1-\lambda)F(\alpha')\\
&\Rightarrow F(\alpha' + \lambda(\alpha - \alpha') ) \leq \lambda F(\alpha)  + (1-\lambda)F(\alpha')\\
&\Rightarrow F(\alpha') + \lambda \langle \alpha - \alpha', \nabla F(\alpha')\rangle + o(\lambda) \leq \lambda F(\alpha)  + (1-\lambda)F(\alpha') \\
&\Rightarrow  \lambda \langle \alpha - \alpha', \nabla F(\alpha')\rangle + o(\lambda) \leq \lambda F(\alpha)  -\lambda F(\alpha') \\
&\Rightarrow  \langle \alpha - \alpha', \nabla F(\alpha')\rangle + \frac{o(\lambda)}{\lambda} \leq  F(\alpha)  - F(\alpha') 
\end{align*}
Letting $\lambda \to 0$ leads to \eqref{seq:convexity}.
\qed
\begin{prop}[Optimality condition]
	Assume F is differentiable and convex on $\cP(\bbR^d)$ then $\alpha^\star$ minimizes F if and only if $\langle \nabla F(\alpha^\star), \alpha - \alpha^\star\rangle \geq 0 $.
\end{prop}
\proof
($\Rightarrow$) Assume $\alpha^\star$ is a minimizer of $F$. Let $t >$. Since $\cP(\bbR^d)$ is convex, we can write for any $\alpha \in \cP(\bbR^d)$:
$$F(\alpha^\star) \leq F(\alpha^\star + t (\alpha - \alpha^\star)) $$
For t small enough, we can use \eqref{seq:differentiability} on the right-hand side:
$$F(\alpha^\star) \leq F(\alpha^\star) +  t\langle \alpha-\alpha^\star, \nabla F(\alpha^\star)\rangle + o(t) $$
Dividing by $t$ and letting $t \to 0$ leads to $\langle \alpha-\alpha^\star, \nabla F(\alpha^\star)\rangle \geq 0$ for all $\alpha$.

($\Leftarrow$) Assume $ \langle \nabla F(\alpha^\star), \alpha - \alpha^\star\rangle \geq 0 $. 
Proposition \ref{sprop:convexity} applies and \eqref{seq:convexity} allows to conclude that $\alpha^\star$ is a minimizer of F.
\qed
\section{Proofs of differentiability and convexity}
\paragraph{Proof of Lemma \ref{lem:convexity}}
\begin{slemma}
	\label{slem:convexity}
	Let $\alpha, \alpha' \in \cG(\bbR^d)$ and let $h_\alpha, h_{\alpha'}$ denote their respective autocorrelation potentials. Then:
	\begin{equation}
	\label{seq:bound-autocorr}
	\int e^{\frac{h_\alpha(x)}{\varepsilon}} \kernel(x, y)  e^{\frac{h_{\alpha'}(y)}{\varepsilon}} \diff \alpha(x) \diff\alpha'(y) \leq 1
	\end{equation}
\end{slemma}
\proof
The left side of \eqref{seq:bound-autocorr} can be equivalently written using Fubini-Tonelli:
\begin{align*}
A = \int &e^{\frac{h_\alpha(x)}{\varepsilon}} \kernel(x, y)  e^{\frac{h_{\alpha'}(y)}{\varepsilon}} \diff \alpha(x) \diff\alpha'(y) \\ &=
\langle e^{\frac{h_\alpha}{\varepsilon}}.\alpha, \cK(e^{\frac{h_{\alpha'}}{\varepsilon}} \alpha')\rangle  \\
&= \langle e^{\frac{h_{\alpha'}}{\varepsilon}}.\alpha', \cK^\top(e^{\frac{h_{\alpha}}{\varepsilon}} \alpha)\rangle \\
&=  \langle e^{\frac{h_{\alpha'}}{\varepsilon}}.\alpha', \cK(e^{\frac{h_{\alpha}}{\varepsilon}} \alpha)\rangle \enspace,
\end{align*}
where the last equality follows from the symmetry of $K$. 
Thus we have:
\begin{equation}
\label{eq:A}
A =  \frac{1}{2} \langle e^{\frac{h_{\alpha'}}{\varepsilon}}.\alpha', \cK(e^{\frac{h_{\alpha}}{\varepsilon}} \alpha)\rangle  +
\frac{1}{2} \langle e^{\frac{h_\alpha}{\varepsilon}}.\alpha, \cK(e^{\frac{h_{\alpha'}}{\varepsilon}} \alpha')\rangle
\end{equation}
Since the optimal transport plans (primal solutions) associated with $\ote^{\otimes}(\alpha, \alpha)$ and $\ote^{\otimes}(\alpha', \alpha')$ integrate to 1, the right side of \eqref{seq:bound-autocorr} can be written:
\begin{equation}
\label{eq:1}
1 =  \frac{1}{2}\langle e^{\frac{h_{\alpha}}{\varepsilon}}.\alpha, \cK(e^{\frac{h_{\alpha}}{\varepsilon}} \alpha)\rangle +
\frac{1}{2}\langle e^{\frac{h_{\alpha'}}{\varepsilon}}.\alpha', \cK(e^{\frac{h_{\alpha'}}{\varepsilon}} \alpha')\rangle 
\end{equation}
Combining \eqref{eq:A} with \eqref{eq:1}, it holds:
\begin{equation*}
1 - A = \frac{1}{2} \langle r, \cK(r)\rangle
\end{equation*}
where $r=e^{\frac{h_{\alpha}}{\varepsilon}}.\alpha - e^{\frac{h_{\alpha'}}{\varepsilon}}.\alpha'$.
Since $K$ is semi-definite positive, $1-A \geq 0$. \qed
\paragraph{Differentiability of $\ote$}
\begin{sprop}
	\label{sprop:gradient}
	Under assumption (1), $\ote$ is differentiable and its gradient is given by:
	\begin{equation}
	\label{seq:gradient}
	\nabla\ote^{\otimes}(\alpha, \beta) = (f, g)
	\end{equation}
	Where $f$ and $g$ satisfy the Sinkhorn fixed point system \eqref{eq:schrodinger} on $\bbR^d$.
\end{sprop}

\proof Consider $\alpha, \beta, \alpha_1, \alpha_2, \beta_1, \beta_2 \in \cG(\bbR^d)$ and denote the displacements $\delta\alpha = \alpha_1 - \alpha_2$ and $\delta\beta = \beta_1 - \beta_2$. Let $\Delta_t$ denote the ratio of \eqref{eq:differentiability}:
\begin{equation}
\label{seq:delta}
\Delta_t = \frac{\ote^{\otimes}(\alpha_t, \beta_t) - \ote^{\otimes}(\alpha, \beta)}{t} \enspace,
\end{equation}
where $\alpha_t = \alpha + t\delta\alpha$ and $\beta_t = \beta + t \delta\beta$.
Similarly to the proof of Proposition 2 of \citet{feydy19}, we derive a lower and upper bound of $\Delta_t$ using suboptimal potentials. On one hand, the pair $(f, g)$ is suboptimal for the dual problem defining $\ote^{\otimes}(\alpha_t, \beta_t)$. Therefore:
\begin{align*}
\ote^{\otimes}(\alpha_t, \beta_t) \geq & \langle \alpha_t, f \rangle + \langle \beta_t, g \rangle \\ &- \varepsilon \langle \alpha_t \otimes \beta_t, \exp\left(\frac{f \oplus g - C}{\varepsilon}\right) \rangle + \varepsilon 
\end{align*}
Therefore, \eqref{eq:otdual} and \eqref{eq:schrodinger} lead to the lower bound:
\begin{equation*}
\Delta_t \geq \langle \delta \alpha, f - \varepsilon\rangle + \langle \delta \beta, g-\varepsilon\rangle + o(1)
\end{equation*}
And similarly we get the upper bound:
\begin{equation*}
\Delta_t \leq \langle \delta \alpha, f_t - \varepsilon\rangle + \langle \delta \beta, g_t-\varepsilon\rangle + o(1)
\end{equation*}
As $t \to 0$, $(\alpha_t, \beta) \to (\alpha, \beta)$. On one hand, Proposition 4 of \citet{mena19} leads to the pointwise convergence of the sequence of potentials $(f_t, g_t)$ towards $(f, g)$. On the other hand, Proposition \ref{prop:mena19} implies that there exists $M > 0$ such that $|f_t(x)| \leq M \|x\|^2$ for all $x \in \bbR^d$. Given that any $\mu \in \cG\sigma(\bbR^d)$ has a second order moment, by Lebesgue's dominated convergence we have $\langle \mu, f_t \rangle \to \langle \mu, f\rangle$. Similarly, $\langle \mu, g_t \rangle \to \langle \mu, g\rangle$. Finally, since $\langle \delta \alpha , \varepsilon\rangle =  \langle \delta \beta , \varepsilon\rangle = 0$, we get as $t \to 0$, $\Delta_t\to\langle \delta\alpha, f\rangle + \langle \delta\beta, g\rangle$. Since $f$ and $g$ are smooth (Prop \ref{prop:mena19}) and square-integrable with respect to any $\mu \in \cG(\bbR^d)$, \eqref{seq:gradient} holds for $\nabla\ote^{\otimes}(\alpha, \beta) = (f, g)$.
\qed

\subsection{Differentiability and convexity of $\ote^{\cL}$}
\label{ss:diff-lebesgue}
To prove theorem \ref{thm:lebesgue} we first need to establish the differentiability and convexity of $\ote^{\cL}$ on the set of sub-Gaussian measures $\cG_{\sigma}(\bbR^d)$ which are absolutely continuous with respect to the Lebesgue measure. 

\paragraph{Dual problem}
Let $\alpha, \beta$ continuous sub-Gaussian measures. Identifying $\alpha$, $\beta$ and $\pi$ with their Lebesgue densities, The OT problem \eqref{eq:otdef} has a dual problem given by:
\begin{equation}
\label{seq:ot-lebesgue-dual}
\ote^\cL(\alpha, \beta) = \sup_{f \in \cL_1(\alpha), g\in\cL_1(\beta)} \langle f, \alpha \rangle + \langle g, \beta \rangle 
-\varepsilon \int\int \exp\left(\frac{f(x) + g(y) - C(x, y)}{\varepsilon}\right) \diff x\diff y + \varepsilon\enspace,
\end{equation}
Notice that the convexity of $\ote^{\cL}$ follows immediately from \eqref{seq:ot-lebesgue-dual}  since it is a supremum of linear functions in $\alpha$ and  $\beta$. The optimality conditions are equivalent to the marginal constraints of the primal problem \eqref{eq:otdef}. However, they are slightly different than those of $\ote^{\otimes}$. Cancelling the gradient of the dual problem leads to the following system \cite{gentil17}:

\begin{align}
\begin{split}
\label{seq:schrodinger-lebesgue}
e^{\frac{f}{\varepsilon}} \cK(e^{\frac{g}{\varepsilon}}) = \alpha\enspace,\\
e^{\frac{g}{\varepsilon}} \cK^\top(e^{\frac{f}{\varepsilon}}) = \beta\enspace,
\end{split}
\end{align}
which in integral form can be written:
\begin{align}
\begin{split}
\label{seq:schrodinger-lebesgue-integral}
e^{\frac{f(x)}{\varepsilon}} \int e^{\frac{- C(x, y) + g(y)}{\varepsilon}} \diff y = \alpha(x) \enspace \forall x,\\
e^{\frac{g(x)}{\varepsilon}} \int e^{\frac{- C(y, x) + f(y)}{\varepsilon}} \diff y = \beta(x) \enspace \forall x,\\
\end{split}
\end{align}

\'and the optimal transport plan's density $\pi$ is given by:
$\pi(x,y)=\exp\left(\frac{f(x) + g(y) - C(x, y)}{\varepsilon}\right)$

Thus, at optimality the integral over $\bbR^{d} \times \bbR^{d}$ sums to 1 and:
\begin{equation}
\label{seq:ot-optimality-lebesgue}
\ote^{\cL}(\alpha, \beta) = \langle f, \alpha \rangle + \langle g, \beta \rangle
\end{equation}

\paragraph{Convexity and Differentiability}
By using the existence of Lebesgue continuity, one can rewrite the $\kl$ in the primal problem such that it holds \citep{dimarino19}:
\begin{align}
\label{seq:reference-l-2}
\ote^\cL(\alpha, \beta) = \ote^\otimes(\alpha, \beta) 
+\varepsilon \kl(\alpha| \cL) + \varepsilon \kl(\beta|\cL) 
\end{align}
We already showed that $\ote^{\otimes}$ is convex (w.r.t. to one argument); $\kl$ is also convex (even jointly convex). Since the set of Lebesgue-continuous and sub-Gaussian measures is convex , $\ote^{\cL}$ is also convex with respect to one argument.

Identifying $\alpha$ with its density, we have $E(\alpha) \eqdef \kl(\alpha, \cL) = \int \alpha(x)(\log(\alpha(x)) - 1) \diff x$. If $\alpha > 0$, then for any  feasible displacement $h = h_1 - h_2$ with density functions $h_1, h_2$. The functional derivative of $E$ in the direction $h$ is given by: $\left[\frac{\diff E(\alpha + th)}{\diff t}\right]_{t=0} = \langle h, \log(\alpha)\rangle$. Thus, in the sense of the directional differentiation \eqref{eq:differentiability}: $\nabla_\alpha \kl(\alpha, \cL) = \log(\alpha)$.

Let $(f, g)$ be a pair of optimal potentials for $\ote^{\otimes}(\alpha, \beta)$. Following \eqref{sprop:gradient} and the differentiability of $\kl$, $\ote^{\cL}$ is differentiable on the set of sub-Gaussian measures with positive density functions and its gradient is given by: $\nabla_1 \ote^{\cL}(\alpha, \beta) = f - \varepsilon\log(\alpha)$. By a simple calculation, it is easy to show that $(f - \varepsilon\log(\alpha), g-\varepsilon\log(\beta)$ are actually solutions of the Sinkhorn equations \eqref{seq:ot-optimality-lebesgue}. Similarly, given a solution $(f_1, g_1)$ of \eqref{seq:ot-lebesgue-dual}, $(f_1 + \varepsilon\log(\alpha), g_1 +  \varepsilon\log(\beta))$ are optimal potentials of $\ote^{\otimes}$. Therefore, the following proposition holds:
\begin{sprop}
	\label{sprop:grad-lebesgue}
	Let $\alpha, \beta \in \cG_{\sigma}(\bbR^d)$. If $\alpha$ and $\beta$ are Lebesgue-continuous with positive density functions, then $\ote^{\cL}$ is differentiable and it holds:
	\begin{equation}
	\label{seq:grad-lebesgue}
	\nabla\ote^{\cL}(\alpha, \beta) = (f, g)\enspace,
	\end{equation}
where $(f, g)$ is a pair of dual potentials verifying the fixed point equations \eqref{seq:ot-optimality-lebesgue}.
	
\end{sprop}

\section{Proofs of the theorems}
\label{ss:gaussians}
We first start by showing that the equations verified by the variance of the barycenter have a unique positive solution.
\subsection{Fixed point equations Lemmas}
For the 3 following lemmas, since the variance $S$ can only be positive, we re-parametrized the equations by replacing $S^2$ with $S$ for the sake of simplicity.
\begin{slemma}
	\label{slem:contraction-lebesgue}
	Under the assumptions of Theorem \ref{thm:lebesgue}, the equation in $S$:
	\begin{equation}
	\label{seq:fx-lebesgue}
	\sum_{k=1} w_k \sqrt{\varepsilon'^4 + 4\sigma_k^2S} = - \varepsilon'^2 + 2S
	\end{equation}
	has a positive solution.
\end{slemma}
\proof
Let $f: S \in \bbR_+ \to \sum_{k=1} w_k \sqrt{\varepsilon'^4 + 4\sigma_k^2S} + \varepsilon'^2 - 2S$. Since $f$ is continuous and $f(0) = 2\varepsilon'^2 > 0$ and $\underset{S\to -\infty}{\lim}f(S) = -\infty$, there exists $S > 0$ such that $f(S) = 0$.
\qed
\begin{slemma}
	\label{slem:contraction-prod}
	Under the assumptions of Theorem \ref{thm:product}, the equation in $S$:
	\begin{equation}
	\label{seq:fx-prod}
	\sum_{k=1} w_k \sqrt{\varepsilon'^4 + 4\sigma_k^2S} = \varepsilon'^2 + 2S
	\end{equation}
	has a positive solution if and only if $\varepsilon'^2 < \bar{\sigma} = \sum_{k=1}{w_k} \sigma_k^{2}$.
\end{slemma}
\proof
Let $f: S \in \bbR_+ \to \sum_{k=1} w_k \sqrt{\varepsilon'^4 + 4\sigma_k^2S} - \varepsilon'^2 - 2S$. 
Sufficient condition. since $f(0) = 0$ and $\underset{S\to +\infty}{\lim}f(S) = -\infty$, a positive solution exists if $f'(0) > 0$.
\begin{align*}
f'(0) > 0 &\Leftrightarrow \sum_{k=1} \frac{w_k \sigma_k^2 }{\varepsilon'^2} - 1 > 0 \\
&\Leftrightarrow \bar{\sigma} > \varepsilon'^2
\end{align*}
Necessary condition. Conversely, we have by Jensen's inequality:
\begin{align*}
f(S) &\leq \sqrt{\varepsilon'^4 + 4\bar{\sigma}S} - \varepsilon'^2 - 2S \\
&= 4S\frac{\bar{\sigma} - S - \varepsilon'^2}{\sqrt{\varepsilon'^4 + 4\bar{\sigma}S} + \varepsilon'^2 + 2S}
\end{align*}
Therefore, if $\bar{\sigma} \leq \varepsilon'^2$ then $f(S) \leq - \frac{4S^2}{\sqrt{\varepsilon'^4 + 4\bar{\sigma}S} + \varepsilon'^2 + 2S} < 0$ for any $S > 0$.
\qed
\begin{slemma}
	\label{slem:contraction-div}
	Under the assumptions of Theorem \ref{thm:div}, the equation in $S$:
	\begin{equation}
	\label{seq:fx-div}
	\sum_{k=1} w_k \sqrt{\varepsilon'^4 + 4\sigma_k^2S} = \sqrt{\varepsilon'^4 + 4S^4}
	\end{equation}
	has a positive solution $S^\star$ and it holds $S^\star \in (\sigma_{(0)}, \sigma_{(K)})$.
\end{slemma}
\proof
Let $f: S \in \bbR_+ \to \sum_{k=1} w_k \sqrt{\varepsilon'^4 + 4\sigma_k^2S} - \sqrt{\varepsilon'^4 + 4S^2}$.  It holds: 
\begin{align*}
f(S) &\geq  \sqrt{\varepsilon'^4 + 4\sigma_{(0)}^2S} - \sqrt{\varepsilon'^4 + 4S^2} \\
& = \frac{4S(\sigma_0^2 - S)}{\sqrt{\varepsilon'^4 + 4\sigma_0^2S} + \sqrt{\varepsilon'^4 + 4S^2} }\enspace.
\end{align*}
Thus $f(\sigma_{(0)}^2) \geq 0$. Similarly $f(\sigma_{(K)}) \leq 0$. Thus there exists $S^\star \in (\sigma_{(0)}, \sigma_{(K)})$ such that $f(S^\star) = 0$.
\qed
\subsection{Proofs of theorems \ref{thm:product} and \ref{thm:div}}
We turn now to proving theorems \ref{thm:div} and \ref{thm:product}.
 We have shown that both $\ote$ and $\sdiv$ are convex and differentiable on the convex set of sub-Gaussian measure on $\bbR^d$. Thus, the proposition holds:
\begin{prop}
	\label{seq:subgaussian-charac}
	Let $\alpha_1, \dots, \alpha_K \in \cG(\bbR^d)$. Let $(w_1, \dots, w_K)$ be non-negative weights summing to 1. Then:
	
	$\alpha_{\ote^{\otimes}} = \argmin_{\alpha} \sum_{k=1}^K w_k \ote^{\otimes}(\alpha_k, \alpha)$ if and only if there exists at set of potentials $f_1, \dots, f_K, g_1, \dots, g_K$ such that for any  direction $\beta \in \cG(\bbR^d)$ the following equations hold everywhere in $\bbR^d$:
	\begin{equation}
	\left\{
	\begin{array}{ll}
	\label{seq:kkt-ote}
	&e^{\frac{f_k}{\varepsilon}}. \cK(e^{\frac{g_k}{\varepsilon}}.  \alpha_{\ote^\otimes}) = 1, \enspace
	e^{\frac{g_k}{\varepsilon}}. \cK^\top(e^{\frac{f_k}{\varepsilon}}.\alpha_k) = 1, \\
	&\langle \sum_{k=1}^K w_k g_k, \beta- \alpha_{\ote^\otimes} \rangle \geq 0
	\end{array}
	\right.
	\end{equation}
	$\alpha_{\sdiv} = \argmin_{\alpha} \sum_{k=1}^K w_k \sdiv(\alpha_k, \alpha)$ if and only if there exists at set of potentials $f_1, \dots, f_k, g_1, \dots, g_k, h$ such that for any direction $\beta \in \cG(\bbR^d)$ the following equations hold everywhere in $\bbR^d$:
	\begin{equation}
	\left\{
	\begin{array}{ll}
	\label{seq:kkt-sdiv}
	&e^{\frac{f_k}{\varepsilon}}. \cK(e^{\frac{g_k}{\varepsilon}}.  \alpha_{\sdiv}) = 1, \enspace
	e^{\frac{g_k}{\varepsilon}}. \cK^\top(e^{\frac{f_k}{\varepsilon}}.\alpha_k) = 1, \\
	&e^{\frac{h}{\varepsilon}}. \cK(e^{\frac{h}{\varepsilon}}.\alpha_{\sdiv}) = 1, 	\\
	&\langle \sum_{k=1}^K w_k g_k - h, \beta- \alpha_{\sdiv} \rangle \geq 0
	\end{array}
	\right.
	\end{equation}
\end{prop}
We solve the systems of equations \eqref{seq:kkt-ote} and \eqref{seq:kkt-sdiv} by restricting the potentials to quadratic functions. Since the objectives are convex, showing the existence of a solution is sufficient for optimality. We start with the Debiased barycenter (theorem \ref{thm:div}).

\begin{stheorem}[Debiasing  of $\sdiv$]
	\label{sthm:div}
	Let $\cost(x, y) = (x - y)^2$ and $0 <~\varepsilon~<~+\infty$ and $\varepsilon = 2\varepsilon'^2$. Let $(w_k)$ be positive weights that sum to 1. Let $\cN$ denote the Gaussian distribution. Assume that $\alpha_k \sim \cN(\mu_k, \sigma_k^2)$  and let $\bar{\mu} = \sum_k w_k \mu_k$, $\bar{\sigma} = \sum_{k=1}{w_k} \sigma_k^{2}$ then:
	
	Then $\alpha_{\sdiv} \sim \cN(\bar{\mu}, S^2) $ where $S$ is the unique non-zero solution $S^\star$ of the fixed point equation: 
	
	$\sum_{k=1} w_k \sqrt{\varepsilon'^4 + 4\sigma_k^2S^2} = \sqrt{\varepsilon'^4 + 4S^4}$. Moreover, given a sorted sequence $\sigma_{(1)} \leq \dots \leq \sigma_{(K)}$, it holds $S^\star \in (\sigma_{(0)}, \sigma_{(K)})$.
	
	In particular, if all $\sigma_k$ are equal to some $\sigma > 0$, then $\alpha_{\sdiv}~\sim~\cN(\bar{\mu}, \sigma^2) $.
\end{stheorem}
\proof
Convexity makes the system equations \eqref{seq:kkt-ote} and \eqref{seq:kkt-sdiv} sufficient for optimality. Thus, we only need to find \emph{a particular} solution. We are going to show that there exist a set of quadratic polynomial potentials and Gaussian probability measures satisfying each system. First, let's start with the $\sdiv$ barycenter $\alpha_{\sdiv}$.
	
Consider polynomial potentials of the form  $f_k(x) = F_{2, k} x^2 + F_{1, k} x + F_{0, k}$ and $g_k(x) = G_{2, k} x^2 + G_{1, k} x + G_{0, k}$ and $h(x) = H_{2} x^2 + H_{1} x + H_{0}$for some unknown coefficients $F_{2, k}, F_{1, k}, F_{0, k},  G_{2, k}, G_{1, k},  G_{0, k}, H_2, H_1, H_0 \in \bbR$, and assume that $\density{\alpha_{\ote}} = \cN(m, S)\enspace$. First, we will write the first and second order coefficients as functions of $m$ and $S$ then use the optimality condition to find $m$ and $S$.

\paragraph{Sufficient optimality condition}
 Let $\beta \in \cG(\bbR^d)$. And let $M_r(\beta)$ denote the r-th moment of $\beta$. For any real sequence $y_1, \dots, y_k$, let $\bar{y}$ denote its weighted average $\sum_{k=1}^K w_k y_k$. The optimality condition reads:
\begin{align*}
&\langle \sum_{k=1}^K w_k g_k - h, \beta- \alpha_{\sdiv} \rangle \geq 0 \\
&\Leftrightarrow  (\bar{G_2} - H_2) (M_2(\beta) - M_2(\alpha_{\sdiv}))  + (\bar{G_1} - H_1) (M_1(\beta) - M_1(\alpha_{\sdiv}))   + (\bar{G_0} - H_0)(M_0(\beta) - M_0(\alpha_{\sdiv})) \geq 0 \\
&\Leftrightarrow  (\bar{G_2} - H_2) (M_2(\beta) - M_2(\alpha_{\sdiv}))  + (\bar{G_1} - H_1) (M_1(\beta) - M_1(\alpha_{\sdiv}))  \geq 0 \\
\end{align*}
Where the last inequality follows from $M_0(\beta) = M_0(\alpha_{\sdiv})=  \int \diff\alpha_{\sdiv} = 1$. Thus, the 0-order coefficients are irrelevant for optimality. We are going to show that there exist a set of coefficients such that the following sufficient conditions for optimality hold:
\begin{align}
\label{eq:optim-condition-coefs}
&\bar{G_2}  \eqdef \sum_{k=1}^K w_k G_{2, k} = H_2  \\
&\bar{G_1}  \eqdef \sum_{k=1}^K w_k G_{1, k} = H_1
\end{align}

\paragraph{Kernel integration}
Dropping the $k$ exponent for the sake of convenience, let's carefully derive the integral $\cK^\top(e^{\frac{f_k}{\varepsilon}}.\alpha_k)$:
	\begin{align*}
	\cK^\top(e^{\frac{f}{\varepsilon}} \alpha_k)(x)  &= \int K(x, y) e^{\frac{f(y)}{2\varepsilon'^2}} \frac{\diff\alpha}{\diff \lambda}(y)  \diff y \\
	 &= \frac{1}{\sqrt{2\pi \sigma^2}} \int \exp\left(\frac{- (x - y)^2 + f(y)}{2\varepsilon'^2} - \frac{(y-\mu)^2} {2\sigma^2} \right) \diff y \\
	 &= \frac{1}{\sqrt{2\pi \sigma^2}} \int \exp\left( \underbrace{\left[\frac{ F_2 - 1}{2\varepsilon'^2} - \frac{1}{2\sigma^2}\right]}_{A} y^2 + \underbrace{\left[\frac{ F_1 }{2\varepsilon'^2} + \frac{x }{\varepsilon'^2} +  \frac{\mu}{\sigma^2}\right]}_{Z(x)} y + \left[\frac{ F_0 - x^2}{2\varepsilon'^2} - \frac{\mu^2}{2\sigma^2}\right] \right) \diff y \\
	 &= \frac{1}{\sqrt{2\pi \sigma^2}} \exp\left(\frac{ F_0 - x^2}{2\varepsilon'^2} - \frac{\mu^2}{2\sigma^2}\right) \int \exp\left( A  \left[y^2 + \frac{Z(x)}{A} y  \right] \right) \diff y \\
	 &= \frac{1}{\sqrt{2\pi \sigma^2}} \exp\left(\frac{ F_0 - x^2}{2\varepsilon'^2} - \frac{\mu^2}{2\sigma^2}\right) \int \exp\left( A  \left[y+ \frac{Z(x)}{2A}  \right]^2 - \frac{Z(x)^2}{4A} \right) \diff y \\
	 &= \frac{1}{\sigma} \exp\left(\frac{ F_0 - x^2}{2\varepsilon'^2} - \frac{\mu^2}{2\sigma^2} - \frac{Z(x)^2}{4A}\right) \underbrace{ \frac{1}{\sqrt{2\pi }}\int \exp\left( A  \left[y+ \frac{Z(x)}{2A} \right]^2  \right) \diff y}_{I} \\
\end{align*}
For the fourth equality to be sound, we need $A \neq 0$, and for the integral $I$ to be finite, we need $A < 0$ which is equivalent to:
\begin{equation}
\label{seq:condition-F2-2}
F_2 < 1 + \frac{\varepsilon'^2}{\sigma^2}.
\end{equation}
In that case, $I = \frac{1}{\sqrt{-2A}}$ thus:
\begin{align*}
\cK^\top(e^{\frac{f}{\varepsilon}} \alpha_k)(x) 	 &= \frac{1}{\sigma\sqrt{-2A}} \exp\left(\frac{ F_0 - x^2}{2\varepsilon'^2} - \frac{\mu^2}{2\sigma^2} - \frac{Z(x)^2}{4A}\right)  \\
&= \frac{1}{\sigma\sqrt{-2A}} \exp\left( \left[-\frac{ 1}{2\varepsilon'^2} - \frac{1}{4A\varepsilon'^4}\right]x^2 - \left[ \frac{F_1}{4A\varepsilon'^4} + \frac{\mu}{2A\sigma^2 \varepsilon'^2} \right]x - \frac{\mu^2}{2\sigma^2} + \frac{F_0}{2\varepsilon'^2} - \frac{\left[\frac{ F_1 }{2\varepsilon'^2} + \frac{\mu}{\sigma^2}\right]^2}{4A} \right)  \\
&=\exp\left( \left[-\frac{ 1}{2\varepsilon'^2} - \frac{1}{4A\varepsilon'^4}\right]x^2 - \left[ \frac{F_1}{4A\varepsilon'^4} + \frac{\mu}{2A\sigma^2 \varepsilon'^2} \right]x - \frac{\mu^2}{2\sigma^2} + \frac{F_0}{2\varepsilon'^2} - \frac{\left[\frac{ F_1 }{2\varepsilon'^2} + \frac{\mu}{\sigma^2}\right]^2}{4A} - \log(\sigma\sqrt{-2A})\right)
\end{align*}
	\paragraph{Sinkhorn equations}
Using the first Sinkhorn equation $e^{\frac{g_k}{\varepsilon}}. \cK^\top(e^{\frac{f_k}{\varepsilon}}.\alpha_k) = 1$ we get by identification, for all $k$, with :
\begin{equation}
\label{seq:defA}
A_k = \frac{F_{2, k} - 1}{2\varepsilon'^2} - \frac{1}{2\sigma_k^2}
\end{equation}
\[
\left\{
\begin{array}{ll}
\frac{G_{2, k} - 1}{\varepsilon'^2} - \frac{1}{2A_k\varepsilon'^4} = 0 \enspace [i] \\
\frac{G_{1, k}}{\varepsilon'^2} - \frac{F_{1, k}}{2A_k\varepsilon'^4} - \frac{\mu_k}{A_k\sigma_k^2\varepsilon'^2} = 0 \enspace[ii] \\
\frac{G_{0, k}}{\varepsilon'^2} - \frac{\mu_k^2}{\sigma_k^2} + \frac{F_{0, k}}{\varepsilon'^2} - \frac{\left[\frac{ F_{1, k} }{2\varepsilon'^2} + \frac{\mu_k}{\sigma_k^2}\right]^2}{2A_k} - \log(-2\sigma_k^2A_k) = 0 \enspace[iii] 
\end{array}
\right.
\]
Similarly, since the equations are symmetric,
 $e^{\frac{f_k}{\varepsilon}}. \cK(e^{\frac{g_k}{\varepsilon}}.\alpha_{\ote}) = 1$ with $G_{2, k} < 1 + \frac{\varepsilon'^2}{S^2}$ and:
 \begin{equation}
 \label{eq:defB}
 B_k = \frac{ G_{2, k} - 1}{2\varepsilon'^2} - \frac{1}{2S^2} 
 \end{equation}lead to:
\[
\left\{
\begin{array}{ll}
\frac{F_{2, k} - 1}{\varepsilon'^2} - \frac{1}{2B_k\varepsilon'^4} = 0 \enspace [j] \\
\frac{F_{1, k}}{\varepsilon'^2} - \frac{G_{1, k}}{2B_k\varepsilon'^4} - \frac{m}{B_k S^2\varepsilon'^2} = 0 \enspace[jj] \\
\frac{F_{0, k}}{\varepsilon'^2} - \frac{m^2}{S^2} + \frac{G_{0, k}}{\varepsilon'^2} - \frac{\left[\frac{ G_{1,k} }{2\varepsilon'^2} + \frac{m}{S^2}\right]^2}{2B_k} - \log(-2S^2B_k) = 0 \enspace[jjj] 
\end{array}
\right.
\]
\paragraph{Second order coefficients $G_2, F_2$}
Let's rewrite [i] and [j] separately:
\begin{align*}
\left\{
\begin{array}{ll}
2B_k + \frac{1}{S^2} - \frac{1}{2A_k\varepsilon'^4} = 0 \enspace  \\
2A_k + \frac{1}{\sigma_k^2} - \frac{1}{2B_k\varepsilon'^4} = 0 \enspace 
\end{array}
\right.
\\
\Leftrightarrow
\left\{
\begin{array}{ll}
	A_k B_k + \frac{A_k}{2S^2} - \frac{1}{4\varepsilon'^4} = 0 \enspace  \\
	A_k B_k + \frac{B_k}{2\sigma_k^2} - \frac{1}{4\varepsilon'^4} = 0 \enspace 
\end{array}
\right.
\\
\Leftrightarrow
\left\{
\begin{array}{ll}
	 \frac{B_k}{\sigma_k^2}  =  \frac{A_k}{S^2}  \enspace \\
	A_k B_k + \frac{B_k}{2\sigma_k^2} - \frac{1}{4\varepsilon'^4} = 0 \enspace 
\end{array}
\right.
\\
\Leftrightarrow
\left\{
\begin{array}{ll}
	\frac{B_k}{\sigma_k^2}  =  \frac{A_k}{S^2}  \enspace \\
	B_k^2 + \frac{B_k}{2S^2} - \frac{\sigma_k^2}{4\varepsilon'^4 S^2} = 0 \enspace 
\end{array}
\right.
\end{align*}
The roots of the polynomial above are: $ - \frac{1}{4S^2} \pm \sqrt{ \frac{1}{16S^4} +  \frac{\sigma_k^2}{4S^2\varepsilon'^4} }$. The constraint $B_k < 0$ eliminates the positive solution and it holds:
\begin{align}
	B_k   &=  - \frac{1}{4S^2} - \sqrt{ \frac{1}{16S^4} +  \frac{\sigma_k^2}{4S^2\varepsilon'^4} } \label{eq:Broot} \enspace \\
	A_k &=  \frac{S^2}{\sigma_k^2}B_k \enspace \label{eq:Aroot}
\end{align}

\paragraph{First order coefficients $G_1, F_1$}

Let's rewrite [ii] and [jj] separately:

\begin{align*}
&\left\{
\begin{array}{ll}
\frac{G_{1, k}}{\varepsilon'^2} - \frac{F_{1, k}}{2A_k\varepsilon'^4} - \frac{\mu_k}{A_k\sigma_k^2\varepsilon'^2} = 0 \enspace[ii] \\
\frac{F_{1, k}}{\varepsilon'^2} - \frac{G_{1, k}}{2B_k\varepsilon'^4} - \frac{m}{B_k S^2\varepsilon'^2} = 0 \enspace[jj] \\
\end{array}
\right.
\\
&\Leftrightarrow
\left\{
\begin{array}{ll}
2A_k G_{1, k} - \frac{F_{1, k}}{\varepsilon'^2} - \frac{2\mu_k}{\sigma_k^2} = 0 \enspace[ii]  \\
\frac{F_{1, k}}{\varepsilon'^2} - \frac{G_{1, k}}{2B_k\varepsilon'^4} - \frac{m}{B_k S^2\varepsilon'^2} = 0 \enspace[jj] \\
\end{array}
\right.
\\
&\Leftrightarrow
\left\{
\begin{array}{ll}
\left(2A_k- \frac{1}{2B_k\varepsilon'^4}\right)G_{1, k}  - \frac{2\mu_k}{\sigma_k^2} - \frac{m}{B_k S^2\varepsilon'^2}= 0 \enspace[ii] + [jj] \\
\frac{F_{1, k}}{\varepsilon'^2} - \frac{G_{1, k}}{2B_k\varepsilon'^4} - \frac{m}{B_k S^2\varepsilon'^2} = 0 \enspace[jj] \\
\end{array}
\right.
\\
\end{align*}
The equations above between $A_k$ and $B_k$ lead to $2A_k- \frac{1}{2B_k\varepsilon'^4} = -\frac{1}{\sigma_k^2}$ and the second order polynomial equation in $B_k$ leads to $\frac{\sigma_k^2}{B_k S^2 \varepsilon'^2} = 4\varepsilon'^2 B_k + 2\varepsilon'^2 \frac{1}{S^2}$. Therefore, [ii] + [jj] can be written:
\begin{align*}
&  \frac{1}{\sigma_k^2} G_{1, k}  + \frac{2\mu_k}{\sigma_k^2} + \frac{m}{B_k S^2\varepsilon'^2} = 0 \enspace \\
&\Rightarrow
G_{1, k}  + 2\mu_k + m(4\varepsilon'^2 B_k + 2\varepsilon'^2 \frac{1}{S^2})= 0 \enspace \\
&\Rightarrow G_{1, k}  + 2\mu_k + 2m (G_{2,k} - 1) = 0
\end{align*}
Using [jj] we recover the first order coefficients $F_{1, k}$ and $G_{1,k}$ as function of $m$:
\begin{align}
\label{eq:first-order-FG}
\begin{split}
G_{1, k}  + 2\mu_k + 2m (G_{2,k} - 1) = 0\\
 F_{1, k}  + 2m + 2\mu_k (F_{2,k} - 1) = 0
 \end{split}
\end{align}

\paragraph{Sinkhorn auto-correlation equation}
Similarly, the auto-correlation equation $e^{\frac{h}{\varepsilon}}. \cK^\top(e^{\frac{h}{\varepsilon}}.\alpha_k) = 1$ leads to the same system of equations (equal dual potentials), with $H_{2} < 1 + \frac{\varepsilon'^2}{S^2}$ and:
\begin{equation}
\label{eq:defC}
C = \frac{H_{2} - 1}{2\varepsilon'^2} - \frac{1}{2S^2} < 0
\end{equation}
\begin{align*}
\left\{
\begin{array}{ll}
\frac{H_{2} - 1}{\varepsilon'^2} - \frac{1}{2C\varepsilon'^4} = 0 \enspace [a] \\
\frac{H_{1}}{\varepsilon'^2} - \frac{H_{1}}{2C\varepsilon'^4} - \frac{\mu}{CS^2\varepsilon'^2} = 0 \enspace[b] \\
\frac{H_{0}}{\varepsilon'^2} - \frac{\mu^2}{S^2} + \frac{H_{0}}{\varepsilon'^2} - \frac{\left[\frac{ H_{1} }{2\varepsilon'^2} + \frac{\mu}{S^2}\right]^2}{2C} - \log(-2S^2C) = 0 \enspace[c] 
\end{array}
\right.
\end{align*}
Isolating [a] we get: $	2C + \frac{1}{S^2}  - \frac{1}{2C\varepsilon'^4} = 0 $.
Again, the only negative root of [a] is given by: 
\begin{equation}
\label{eq:Croot}C = - \frac{1}{4S^2} - \sqrt{ \frac{1}{16S^4} +  \frac{1}{4\varepsilon'^4} } \enspace
\end{equation}
and similarly to \eqref{eq:first-order-FG}, we also get the link between $H_1$ and $H_2$:
\begin{align}
\label{eq:first-order-H}
\begin{split}
H_{1} + 2m H_{2} = 0
\end{split}
\end{align}

\paragraph{Optimality condition and identifying $\sigma$ and $\mu$}
Using the definition of $B_2$ \eqref{eq:defB} and \eqref{eq:defC} and then with their closed form formulas \eqref{eq:Broot} and \eqref{eq:Croot}, the first sufficient optimality condition \eqref{eq:optim-condition-coefs} reads:
\begin{align*}
\sum_{k=1}^K w_k G_{2, k} = H_2& \Rightarrow \sum_{k=1}^K w_k B_{2,k} = \frac{H_2-1}{2\varepsilon'^2} - \frac{1}{2S^2}\\
&\Rightarrow \sum_{k=1}^K w_k B_{2,k} = C \\
&\Rightarrow \sum_{k=1}^K w_k \left( \frac{1}{4S^2}  + \sqrt{ \frac{1}{16S^4} +  \frac{\sigma_k^2}{4S^2\varepsilon'^4} } \right) = \frac{1}{4S^2} + \sqrt{ \frac{1}{16S^4} +  \frac{1}{4\varepsilon'^4} } \\
&\Rightarrow \sum_{k=1}^K w_k  \sqrt{ \frac{1}{16S^4} +  \frac{\sigma_k^2}{4S^2\varepsilon'^4} }  = \sqrt{ \frac{1}{16S^4} +  \frac{1}{4\varepsilon'^4} } \\
&\Rightarrow \sum_{k=1}^K w_k  \sqrt{4\sigma_k^2S^2+  \varepsilon'^4} = \sqrt{ 4 S^4 + \varepsilon'^4}
\end{align*} 
Lemma \ref{slem:contraction-div} guarantees that the fixed point equation above possesses a unique positive solution $S$. 

The second sufficient optimality condition \eqref{eq:optim-condition-coefs} combined with the equations on $G_1, F_1$ \eqref{eq:first-order-FG} and $H_1$ \eqref{eq:first-order-H} lead to identifying $m$:
\begin{align*}
\sum_{k=1}^K w_k G_{1, k} = H_1& \Rightarrow m = \sum_{k=1}^K w_k \mu_k
\end{align*} 

\paragraph{Identifying the offset coefficients $F_0, G_0, H_0$}
Since now $m$ and $S$ are known and unique, all the first and second order coefficients $F_{2, k}, G_{2, k}, H_2, F_{1, k}, G_{1, k}, H_1$ are uniquely determined. $H_0$ follows immediately from [c]. Finding $F_{0, k}$ and $G_{0, k}$ can be done up to an additive constant. Adding [iii] and [jjj] leads to a closed form expression on $F_{0, k} + G_{0, k}$. Since the optimality condition does not depend on $H_0, F_0$ and $G_0$,  one may simply set $F_{0, k}$ to 0, and solve $G_{0, k}$ exactly. 
\qed

\begin{stheorem}[Shrinking bias of $\ote^{\otimes}$]
	\label{sthm:product}
	Let $\cost(x, y) = (x - y)^2$ and $0 <~\varepsilon~<~+\infty$ and $\varepsilon = 2\varepsilon'^2$. Let $(w_k)$ be positive weights that sum to 1. Let $\cN$ denote the Gaussian distribution. Assume that $\alpha_k \sim \cN(\mu_k, \sigma_k^2)$  and let $\bar{\mu} = \sum_k w_k \mu_k$, $\bar{\sigma} = \sum_{k=1}{w_k} \sigma_k^{2}$ then:
	
	if $\varepsilon'^2 < \bar{\sigma}$ then $\alpha_{\ote^{\otimes}} \sim \cN(\bar{\mu}, S^2) $ where $S$ is the unique non-zero solution of the fixed point equation: $\sum_{k=1} w_k \sqrt{\varepsilon'^4 + 4\sigma_k^2S^2} = \varepsilon'^2 + 2S^2$. In particular, if all $\sigma_k$ are equal to some $\sigma > 0$, then
	$\alpha_{\ote^{\otimes}}~\sim~\cN(\bar{\mu}, \sigma^2 - \varepsilon'^2) $.
	
	if $\varepsilon'^2 \geq \bar{\sigma}$ then $\alpha_{\ote^{\otimes}}$ is a Dirac distribution located at $\bar{\mu}$.
\end{stheorem}
\proof
When $\varepsilon'^2 < \bar{\sigma}$, the same proof of theorem \ref{sthm:div} applies. Convexity makes the system equations \eqref{seq:kkt-ote} sufficient for optimality. Thus, we only need to find \emph{a particular} solution. We are going to show that there exist a set of quadratic polynomial potentials and Gaussian probability measures satisfying each system. 

Consider polynomial potentials of the form  $f_k(x) = F_{2, k} x^2 + F_{1, k} x + F_{0, k}$ and $g_k(x) = G_{2, k} x^2 + G_{1, k} x + G_{0, k}$for some unknown coefficients $F_{2, k}, F_{1, k}, F_{0, k},  G_{2, k}, G_{1, k},  G_{0, k} \in \bbR$, and assume that $\density{\alpha_{\ote}} = \cN(m, S)\enspace$. First, we will write the first and second order coefficients as functions of $m$ and $S$ then use the optimality condition to find $m$ and $S$.

\paragraph{Sufficient optimality condition}
Let $\beta \in \cP_2(\bbR^d)$. And let $M_r(\beta)$ denote the r-th moment of $\beta$. For any real sequence $y_1, \dots, y_k$, let $\bar{y}$ denote its weighted average $\sum_{k=1}^K w_k y_k$. The optimality condition reads:
\begin{align*}
&\langle \sum_{k=1}^K w_k g_k , \beta- \alpha_{\sdiv} \rangle \geq 0 \\
&\Leftrightarrow  \bar{G_2}  (M_2(\beta) - M_2(\alpha_{\sdiv}))  + \bar{G_1}  (M_1(\beta) - M_1(\alpha_{\sdiv}))   + \bar{G_0} (M_0(\beta) - M_0(\alpha_{\sdiv})) \geq 0 \\
&\Leftrightarrow  \bar{G_2}  (M_2(\beta) - M_2(\alpha_{\sdiv}))  + \bar{G_1} (M_1(\beta) - M_1(\alpha_{\sdiv}))  \geq 0 \\
\end{align*}
Where the last inequality follows from $M_0(\beta) = M_0(\alpha_{\sdiv})=  \int \diff\alpha_{\sdiv} = 1$. Thus, the 0-order coefficients are irrelevant for optimality. 

\paragraph{1. Case 1: if $\varepsilon'^2 < \bar{\sigma}$:}
We are going to show that there exist a set of coefficients such that the following sufficient conditions for optimality hold:
\begin{align}
\label{eq:optim-condition-coefs-ote}
&\bar{G_2}  \eqdef \sum_{k=1}^K w_k G_{2, k} = 0  \\
&\bar{G_1}  \eqdef \sum_{k=1}^K w_k G_{1, k} = 0
\end{align}

The Sinkhorn system on $f_k$ and $g_k$ is the same as in the proof above. Thus, the same equations still hold. The first differences arise when using the optimality condition \eqref{eq:optim-condition-coefs-ote}:

\paragraph{Optimality condition and identifying $\sigma$ and $\mu$}
Using the definition of $B_2$ \eqref{eq:defB}  and its closed form formulas \eqref{eq:Broot}, the first sufficient optimality condition \eqref{eq:optim-condition-coefs-ote} reads:
\begin{align*}
\sum_{k=1}^K w_k G_{2, k} = 0& \Rightarrow \sum_{k=1}^K w_k B_{2,k} = -\frac{1}{2\varepsilon'^2} - \frac{1}{2S^2}\\
&\Rightarrow \sum_{k=1}^K w_k \left( \frac{1}{4S^2}  + \sqrt{ \frac{1}{16S^4} +  \frac{\sigma_k^2}{4S^2\varepsilon'^4} } \right) = \frac{1}{2\varepsilon'^2} + \frac{1}{2S^2} \\
&\Rightarrow \sum_{k=1}^K w_k  \sqrt{4\sigma_k^2S^2+  \varepsilon'^4}  = 2S^2 +  \varepsilon'^2 \\
\end{align*} 
Lemma \ref{slem:contraction-prod} guarantees that the fixed point equation above possesses a unique positive solution $S$ when $\varepsilon'^2 < \bar{\sigma} = \sum_{k=1}w_k \sigma_k^2$.

The second sufficient optimality condition \eqref{eq:optim-condition-coefs-ote} combined with the equations on $G_1, F_1$ \eqref{eq:first-order-FG} lead to identifying $m$:
\begin{align*}
\sum_{k=1}^K w_k G_{1, k} = 0 & \Rightarrow m = \sum_{k=1}^K w_k \mu_k
\end{align*} 

\paragraph{Identifying the offset coefficients $F_0, G_0$}
Since now $m$ and $S$ are known and unique, all the first and second order coefficients $F_{2, k}, G_{2, k}, F_{1, k}, G_{1, k}$ are uniquely determined. Finding $F_{0, k}$ and $G_{0, k}$ can be done up to an additive constant. Adding [iii] and [jjj] leads to a closed form expression on $F_{0, k} + G_{0, k}$. Since the optimality condition does not depend on $H_0, F_0$ and $G_0$,  one may simply set $F_{0, k}$ to 0, and solve $G_{0, k}$ exactly. 

\paragraph{2. Case 2: if $\varepsilon'^2 \geq \bar{\sigma}$:}
We are going to show that there exist a set of potentials such that the Dirac at $\bar{\mu} = \sum_{k=1}w_k \mu_k$ verifies the optimality conditions \eqref{seq:kkt-ote}. Let's simplify the optimality condition for a Dirac minimizer $\alpha_{\sdiv} = \delta_{\bar{\mu}}$

\begin{align*}
&\langle \sum_{k=1}^K w_k g_k , \beta- \alpha_{\sdiv} \rangle \geq 0 \\
&\Leftrightarrow  \bar{G_2}  M_2(\beta) + \bar{G_1}  M_1(\beta)  - \bar{G_2} \bar{\mu}^2 - \bar{G_1} \bar{\mu} \geq 0
\end{align*}
However since for any measure $\beta, M_1(\beta)^2 \leq M_2(\beta)$, the following condition is sufficient for optimality:
\begin{equation}
\label{eq:optim-dirac}
 (\forall x \in \bbR)  \quad \bar{G_2}  x^2 + \bar{G_1}  x - \bar{G_2} \bar{\mu}^2 - \bar{G_1} \bar{\mu} \geq 0
\end{equation}
\paragraph{Sinkhorn equations}
Using the second Sinkhorn equation with $\alpha_{\sdiv} = \delta_{\bar{\mu}}$ given by $e^{\frac{f_k}{\varepsilon}}. \cK(e^{\frac{g_k}{\varepsilon}}.\alpha_{\ote}) = 1$:
\[
\left\{
\begin{array}{ll}
F_{2, k} - 1 = 0 \enspace [j] \\
F_{1, k} + 2m = 0 \enspace[jj] \\
F_{0, k}  - \bar{\mu}^2 + G_{2, k}\bar{\mu}^2 +  G_{1,k}\bar{\mu} + G_{0, k} = 0 \enspace[jjj] 
\end{array}
\right.
\]
Using the first Sinkhorn equation $e^{\frac{g_k}{\varepsilon}}. \cK^\top(e^{\frac{f_k}{\varepsilon}}.\alpha_k) = 1$ we get by identification, for all $k$, with :
\begin{equation}
\label{seq:defA-ote}
A_k = \frac{F_{2, k} - 1}{2\varepsilon'^2} - \frac{1}{2\sigma_k^2}
\end{equation}
\[
\left\{
\begin{array}{ll}
\frac{G_{2, k} - 1}{\varepsilon'^2} - \frac{1}{2A_k\varepsilon'^4} = 0 \enspace [i] \\
\frac{G_{1, k}}{\varepsilon'^2} - \frac{F_{1, k}}{2A_k\varepsilon'^4} - \frac{\mu_k}{A_k\sigma_k^2\varepsilon'^2} = 0 \enspace[ii] \\
\frac{G_{0, k}}{\varepsilon'^2} - \frac{\mu_k^2}{\sigma_k^2} + \frac{F_{0, k}}{\varepsilon'^2} - \frac{\left[\frac{ F_{1, k} }{2\varepsilon'^2} + \frac{\mu_k}{\sigma_k^2}\right]^2}{2A_k} - \log(-2\sigma_k^2A_k) = 0 \enspace[iii] 
\end{array}
\right.
\]
Combining both systems:
\[
\left\{
\begin{array}{ll}
F_{2, k} = 1 \enspace [j] \\
F_{1, k} =- 2\bar{\mu} \enspace[jj] \\
G_{2, k}  = 1 - \frac{\sigma_k^2}{\varepsilon'^2} \enspace [i] \\
G_{1, k} = \frac{2\bar{\mu}\sigma_k^2}{\varepsilon'^2} - 2\mu_k\enspace[ii] \\
\frac{G_{0, k}}{\varepsilon'^2} - \frac{\mu_k^2}{\sigma_k^2} + \frac{F_{0, k}}{\varepsilon'^2} + \sigma_k^2 \left[\frac{ \bar{\mu} }{\varepsilon'^2} - \frac{\mu_k}{\sigma_k^2}\right]^2 = 0 \enspace[iii] 
\end{array}
\right.
\\
\Rightarrow
\left\{
\begin{array}{ll} 
\bar{G_{2}} = 1 - \frac{\bar{\sigma}}{\varepsilon'^2}\\
\bar{G_{1}} = 2\bar{\mu} \left(\frac{\bar{\sigma}}{\varepsilon'^2} - 1\right)
\end{array}
\right.
\]
[iii] can be simplified, it leads to: $ G_{0, k} + F_{0, k} + \frac{\sigma_k^2 \bar{\mu}^2}{\varepsilon'^2} - 2 \bar{\mu}\mu_k = 0$. Similarly, [jjj] can be written: $ F_{0, k}   + G_{0, k} -  \bar{\mu}^2 \frac{\sigma_k^2}{\varepsilon'^2} +  \left(\frac{2\bar{\mu}\sigma_k^2}{\varepsilon'^2} - 2\mu_k\right)\bar{\mu} = 0$ which are equivalent.

Using the assumption $\varepsilon'^2 > \bar{\sigma}$, the optimality condition \eqref{eq:optim-dirac} is equivalent to:
\begin{align*}
 (\forall x \in \bbR) & \quad \bar{G_2}  x^2 + \bar{G_1}  x - \bar{G_2} \bar{\mu}^2 - \bar{G_1} \bar{\mu} \geq 0 \\ 
 \Leftrightarrow (\forall x \in \bbR) & \quad \left(1 - \frac{\bar{\sigma}}{\varepsilon'^2}\right) x^2 + 2\bar{\mu} \left(\frac{\bar{\sigma}}{\varepsilon'^2} - 1\right) x - \left(1 - \frac{\bar{\sigma}}{\varepsilon'^2}\right) \bar{\mu}^2 - 2\bar{\mu} \left(\frac{\bar{\sigma}}{\varepsilon'^2} - 1\right)\bar{\mu} \geq 0 \\
  \Leftrightarrow (\forall x \in \bbR) & \quad x^2 - 2\bar{\mu}  x + \bar{\mu}^2 \geq 0 \\
  \Leftrightarrow (\forall x \in \bbR)  &\quad (x - \bar{\mu})^2 \geq 0 
\end{align*}
Thus, the optimality condition holds.

As before, $F_{0, k}, G_{0, k}$ can be determined up to a constant using [iii]. \qed

\subsection{Proof of theorem \ref{thm:lebesgue}}
\label{ss:proof-lebesgue}
We showed in the supplementary section \ref{ss:diff-lebesgue} that $\ote^{\cL}$ is convex and differentiable on the set of sub-Gaussian measures with positive densities with respect to the Lebesgue measure. Thus, the barycenter $\alpha_{\ote^\cL}$ can be characterized by the first order optimality condition:
\begin{sprop}
	\label{sprop:subgaussian-charac-lebesgue}
	Let $\alpha_1, \dots, \alpha_K \in \cG(\bbR^d)$ be Lebesgue-continuous measures with positive density functions. Let $(w_1, \dots, w_K)$ be non-negative weights summing to 1. Then:
	
	$\alpha_{\ote^{\cL}} = \argmin_{\alpha} \sum_{k=1}^K w_k \ote^{\cL}(\alpha_k, \alpha)$ if and only if there exists at set of potentials $f_1, \dots, f_K, g_1, \dots, g_K$ such that for any  feasible (Lebesgue-continuous) direction $\beta \in \cG(\bbR^d)$ the following equations hold everywhere in $\bbR^d$, identifying the measures with their density functions:
	\begin{equation}
	\left\{
	\begin{array}{ll}
	\label{seq:kkt-otl}
	&e^{\frac{f}{\varepsilon}} \cK(e^{\frac{g}{\varepsilon}} )=   \alpha_k,  \enspace
	e^{\frac{g}{\varepsilon}} \cK^\top(e^{\frac{f}{\varepsilon}} )=\alpha_{\ote^\cL} , \\
	&\langle \sum_{k=1}^K w_k g_k, \beta- \alpha_{\ote^\cL} \rangle \geq 0
	\end{array}
	\right.
	\end{equation}
\end{sprop}

\begin{stheorem}[Blurring bias of $\ote^{\cL}$]
	\label{sthm:lebesgue}
	Let $\cost(x, y) = (x - y)^2$ and $0 <~\varepsilon~<~+\infty$ and $\varepsilon = 2\varepsilon'^2$. Let $(w_k)$ be positive weights that sum to 1. Let $\cN$ denote the Gaussian distribution. Assume $\alpha_k \sim \cN(\mu_k, \sigma_k^2)$  and let $\bar{\mu} = \sum_k w_k \mu_k$,
	
	then $\alpha_{\ote^{\cL}}  \sim \cN(\bar{\mu}, S^2) $ where $S$ is the unique solution of the fixed point equation: 
	
	$\sum_{k=1} w_k \sqrt{\varepsilon'^4 + 4\sigma_k^2S^2} = - \varepsilon'^2 + 2S^2$. 
	
	In particular, if all $\sigma_k$ are equal to some $\sigma > 0$, then
	
	then $\alpha_{\ote^{\cL}} \sim \cN(\bar{\mu}, \sigma^2 + \varepsilon'^2) $.
\end{stheorem}
\proof
The proof of this theorem is technically identical to that of \ref{thm:div}. Except that the Sinkhorn-equations are slightly different.

Consider polynomial potentials of the form  $f_k(x) = F_{2, k} x^2 + F_{1, k} x + F_{0, k}$ and $g_k(x) = G_{2, k} x^2 + G_{1, k} x + G_{0, k}$for some unknown coefficients $F_{2, k}, F_{1, k}, F_{0, k},  G_{2, k}, G_{1, k},  G_{0, k} \in \bbR$, and assume that $\density{\alpha_{\cL}} = \cN(m, S)\enspace$. First, we will write the first and second order coefficients as functions of $m$ and $S$ then use the optimality condition to find $m$ and $S$.

\paragraph{Sufficient optimality condition}
Let a continuous measure $\beta \in \cG(\bbR^d)$ identified with a positive density function. And let $M_r(\beta)$ denote the r-th moment of $\beta$. For any real sequence $y_1, \dots, y_k$, let $\bar{y}$ denote its weighted average $\sum_{k=1}^K w_k y_k$. The optimality condition reads:
\begin{align*}
&\langle \sum_{k=1}^K w_k g_k, \beta- \alpha_{\cL} \rangle \geq 0 \\
&\Leftrightarrow  \bar{G_2}  (M_2(\beta) - M_2(\alpha_{\cL}))  + \bar{G_1}  (M_1(\beta) - M_1(\alpha_{\cL}))   + \bar{G_0}(M_0(\beta) - M_0(\alpha_{\cL})) \geq 0 \\
&\Leftrightarrow  \bar{G_2} (M_2(\beta) - M_2(\alpha_{\cL}))  + \bar{G_1} (M_1(\beta) - M_1(\alpha_{\cL}))  \geq 0 \\
\end{align*}
Where the last inequality follows from $M_0(\beta) = M_0(\alpha_{\cL})=  \int \diff\alpha_{\cL} = 1$. Thus, the 0-order coefficients are irrelevant for optimality. We are going to show that there exist a set of coefficients such that the following sufficient conditions for optimality hold:
\begin{align}
\label{seq:optim-condition-lebesgue}
&\bar{G_2}  \eqdef \sum_{k=1}^K w_k G_{2, k}   \\
&\bar{G_1}  \eqdef \sum_{k=1}^K w_k G_{1, k}
\end{align}

\paragraph{Kernel integration}
Dropping the $k$ exponent for the sake of convenience, let's carefully derive the integral $\cK^\top(e^{\frac{f_k}{\varepsilon}})$:
\begin{align*}
\cK^\top e^{\frac{f}{\varepsilon}}   &= \int K(x, y) e^{\frac{f(y)}{2\varepsilon'^2}} \diff y \\
	 &= \int \exp\left(\frac{- (x - y)^2 + f(y)}{2\varepsilon'^2}  \right) \diff y \\
&=  \int \exp\left( \underbrace{\left[\frac{ F_2 - 1}{2\varepsilon'^2} \right]}_{A} y^2 + \underbrace{\left[\frac{ F_1 }{2\varepsilon'^2} + \frac{x }{\varepsilon'^2} \right]}_{Z(x)} y + \left[\frac{ F_0 - x^2}{2\varepsilon'^2} \right] \right) \diff y \\
&=  \exp\left(\frac{ F_0 - x^2}{2\varepsilon'^2} \right) \int \exp\left( A  \left[y^2 + \frac{Z(x)}{A} y  \right] \right) \diff y \\
&= \exp\left(\frac{ F_0 - x^2}{2\varepsilon'^2} \right) \int \exp\left( A  \left[y+ \frac{Z(x)}{2A}  \right]^2 - \frac{Z(x)^2}{4A} \right) \diff y \\
&= \exp\left(\frac{ F_0 - x^2}{2\varepsilon'^2}- \frac{Z(x)^2}{4A}\right) \underbrace{ \int \exp\left( A  \left[y+ \frac{Z(x)}{2A} \right]^2  \right) \diff y}_{I} \\
\end{align*}
For the fourth equality to be sound, we need $A \neq 0$, and for the integral $I$ to be finite, we need $A < 0$ which is equivalent to:
\begin{equation}
\label{seq:condition-F2-1}
F_2 < 1
\end{equation}
In that case, $I = \sqrt{-\frac{\pi}{A}}$ thus:
\begin{align*}
\cK^\top(e^{\frac{f}{\varepsilon}} )(x) 	 &= \sqrt{-\frac{\pi}{A}}\exp\left(\frac{ F_0 - x^2}{2\varepsilon'^2}  - \frac{Z(x)^2}{4A}\right)  \\
&=\sqrt{-\frac{\pi}{A}}\exp\left( \left[-\frac{ 1}{2\varepsilon'^2} - \frac{1}{4A\varepsilon'^4}\right]x^2 - \left[ \frac{F_1}{4A\varepsilon'^4}  \right]x + \frac{F_0}{2\varepsilon'^2} - \frac{\left[\frac{ F_1 }{2\varepsilon'^2} \right]^2}{4A} \right)  \\
&=\exp\left( \left[-\frac{ 1}{2\varepsilon'^2} - \frac{1}{4A\varepsilon'^4}\right]x^2 - \left[ \frac{F_1}{4A\varepsilon'^4} \right]x  + \frac{F_0}{2\varepsilon'^2} - \frac{\left[\frac{ F_1 }{2\varepsilon'^2} \right]^2}{4A} - \log(\sqrt{-\frac{A}{\pi}})\right)
\end{align*}
\paragraph{Sinkhorn equations}
Using the first Sinkhorn equation $e^{\frac{g_k}{\varepsilon}} \cK^\top(e^{\frac{f_k}{\varepsilon}}) = \alpha_{\cL}$ we get by identification, for all $k$, with :
\begin{equation}
\label{eq:defA-l}
A_k = \frac{F_{2, k} - 1}{2\varepsilon'^2} 
\end{equation}
\[
\left\{
\begin{array}{ll}
\frac{G_{2, k} - 1}{\varepsilon'^2} - \frac{1}{2A_k\varepsilon'^4} +\frac{1}{S^2}= 0 \enspace [i] \\
\frac{G_{1, k}}{\varepsilon'^2} - \frac{F_{1, k}}{2A_k\varepsilon'^4} - \frac{2m}{S^2} = 0 \enspace[ii] \\
\frac{G_{0, k}}{\varepsilon'^2} + \frac{m^2}{S^2} + \frac{F_{0, k}}{\varepsilon'^2} - \frac{\left[\frac{ F_{1, k} }{2\varepsilon'^2} \right]^2}{2A_k} - \log(-\frac{A_k}{\pi}) + \log(2\pi S^2) = 0 \enspace[iii] 
\end{array}
\right.
\]
Similarly, since the equations are symmetric,
$e^{\frac{f_k}{\varepsilon}}. \cK e^{\frac{g_k}{\varepsilon}} = \alpha_{k}) $ with $G_{2, k} < 1 $ and:
\begin{equation}
\label{eq:defB-l}
B_k = \frac{ G_{2, k} - 1}{2\varepsilon'^2} 
\end{equation}lead to:
\[
\left\{
\begin{array}{ll}
\frac{F_{2, k} - 1}{\varepsilon'^2} - \frac{1}{2B_k\varepsilon'^4} +\frac{1}{\sigma_k^2}= 0 \enspace [i] \\
\frac{F_{1, k}}{\varepsilon'^2} - \frac{G_{1, k}}{2B_k\varepsilon'^4} - \frac{2\mu_k}{\sigma_k^2} = 0 \enspace[ii] \\
\frac{F_{0, k}}{\varepsilon'^2} + \frac{\mu_k^2}{\sigma_k^2} + \frac{G_{0, k}}{\varepsilon'^2} - \frac{\left[\frac{ G_{1, k} }{2\varepsilon'^2} \right]^2}{2B_k} - \log(-\frac{B_k}{\pi}) + \log(2\pi \sigma_k^2) = 0 \enspace[iii] 
\end{array}
\right.
\]
\paragraph{Second order coefficients $G_2, F_2$}
Let's rewrite [i] and [j] separately:
\begin{align*}
\left\{
\begin{array}{ll}
2B_k + \frac{1}{S^2} - \frac{1}{2A_k\varepsilon'^4} = 0 \enspace  \\
2A_k + \frac{1}{\sigma_k^2} - \frac{1}{2B_k\varepsilon'^4} = 0 \enspace 
\end{array}
\right.
\\
\Leftrightarrow
\left\{
\begin{array}{ll}
A_k B_k + \frac{A_k}{2S^2} - \frac{1}{4\varepsilon'^4} = 0 \enspace  \\
A_k B_k + \frac{B_k}{2\sigma_k^2} - \frac{1}{4\varepsilon'^4} = 0 \enspace 
\end{array}
\right.
\\
\Leftrightarrow
\left\{
\begin{array}{ll}
\frac{B_k}{\sigma_k^2}  =  \frac{A_k}{S^2}  \enspace \\
A_k B_k + \frac{B_k}{2\sigma_k^2} - \frac{1}{4\varepsilon'^4} = 0 \enspace 
\end{array}
\right.
\\
\Leftrightarrow
\left\{
\begin{array}{ll}
\frac{B_k}{\sigma_k^2}  =  \frac{A_k}{S^2}  \enspace \\
B_k^2 + \frac{B_k}{2S^2} - \frac{\sigma_k^2}{4\varepsilon'^4 S^2} = 0 \enspace 
\end{array}
\right.
\end{align*}
The roots of the polynomial above are: $ - \frac{1}{4S^2} \pm \sqrt{ \frac{1}{16S^4} +  \frac{\sigma_k^2}{4S^2\varepsilon'^4} }$. The constraint $B_k < 0$ eliminates the positive solution and it holds:
\begin{align}
B_k   &=  - \frac{1}{4S^2} - \sqrt{ \frac{1}{16S^4} +  \frac{\sigma_k^2}{4S^2\varepsilon'^4} } \label{eq:Broot-l} \enspace \\
A_k &=  \frac{S^2}{\sigma_k^2}B_k \enspace \label{eq:Aroot-l}
\end{align}

\paragraph{First order coefficients $G_1, F_1$}

Let's rewrite [ii] and [jj] separately:

\begin{align*}
&\left\{
\begin{array}{ll}
\frac{G_{1, k}}{\varepsilon'^2} - \frac{F_{1, k}}{2A_k\varepsilon'^4} - \frac{2m}{S^2} = 0 \enspace[ii] \\
\frac{F_{1, k}}{\varepsilon'^2} - \frac{G_{1, k}}{2B_k\varepsilon'^4} - \frac{2\mu_k}{\sigma_k^2} = 0 \enspace[jj] \\
\end{array}
\right.
\\
&\Leftrightarrow
\left\{
\begin{array}{ll}
2A_k G_{1, k} - \frac{F_{1, k}}{\varepsilon'^2}  - \frac{4A_k\varepsilon'^2 m}{S^2} = 0 \enspace[ii]  \\
\frac{F_{1, k}}{\varepsilon'^2} - \frac{G_{1, k}}{2B_k\varepsilon'^4} - \frac{2\mu_k}{\sigma_k^2} = 0 \enspace[jj] \\
\end{array}
\right.
\\
&\Leftrightarrow
\left\{
\begin{array}{ll}
\left(2A_k- \frac{1}{2B_k\varepsilon'^4}\right)G_{1, k} - \frac{4A_k\varepsilon'^2 m}{S^2} - \frac{2\mu_k}{\sigma_k^2}= 0 \enspace[ii] + [jj] \\
\frac{F_{1, k}}{\varepsilon'^2} - \frac{G_{1, k}}{2B_k\varepsilon'^4} - \frac{2\mu_k}{\sigma_k^2} = 0 \enspace[jj] \\
\end{array}
\right.
\\
\end{align*}
The equations above between $A_k$ and $B_k$ lead to $2A_k- \frac{1}{2B_k\varepsilon'^4} = -\frac{1}{\sigma_k^2}$ and  $A_k = \frac{S^2}{\sigma_k^2}B_k$ leads to: 
\begin{align*}
&  \frac{1}{\sigma_k^2} G_{1, k}  + \frac{2\mu_k}{\sigma_k^2} + \frac{4B_k\varepsilon'^2m}{\sigma_k^2} = 0 \enspace \\
&\Rightarrow
G_{1, k}  + 2\mu_k + 2m(G_{2, k} - 1) = 0 \enspace 
\end{align*}
Using [jj] we recover the first order coefficients $F_{1, k}$ and $G_{1,k}$ as function of $m$:
\begin{align}
\label{eq:first-order-FG-l}
\begin{split}
G_{1, k}  + 2\mu_k + 2m (G_{2,k} - 1) = 0\\
F_{1, k}  + 2m + 2\mu_k (F_{2,k} - 1) = 0
\end{split}
\end{align}

\paragraph{Optimality condition and identifying $\sigma$ and $\mu$}
Using the definition of $B_2$ \eqref{eq:defB-l} and then with their closed form formulas \eqref{eq:Broot-l}, the first sufficient optimality condition \eqref{seq:optim-condition-lebesgue} reads:
\begin{align*}
\sum_{k=1}^K w_k G_{2, k} = 0& \Rightarrow \sum_{k=1}^K w_k B_{2,k} = -\frac{1}{2\varepsilon'^2}\\
&\Rightarrow \sum_{k=1}^K w_k \left( \frac{1}{4S^2}  + \sqrt{ \frac{1}{16S^4} +  \frac{\sigma_k^2}{4S^2\varepsilon'^4} } \right) = \frac{1}{2\varepsilon^2}  \\
&\Rightarrow \sum_{k=1}^K w_k  \sqrt{ \frac{1}{16S^4} +  \frac{\sigma_k^2}{4S^2\varepsilon'^4} }  =  \frac{1}{2\varepsilon^2}  - \frac{1}{4S^2}\\
&\Rightarrow \sum_{k=1}^K w_k  \sqrt{4\sigma_k^2S^2+  \varepsilon'^4} = 2S^2 - \varepsilon'^2
\end{align*} 
Lemma \ref{slem:contraction-lebesgue} guarantees that the fixed point equation above possesses a unique positive solution $S$. 

The second sufficient optimality condition \eqref{seq:optim-condition-lebesgue} combined with the equations on $G_1, F_1$ \eqref{eq:first-order-FG-l} lead to identifying $m$:
\begin{align*}
\sum_{k=1}^K w_k G_{1, k} = 0& \Rightarrow m = \sum_{k=1}^K w_k \mu_k
\end{align*} 

\paragraph{Identifying the offset coefficients $F_0, G_0$}
Since now $m$ and $S$ are known and unique, all the first and second order coefficients $F_{2, k}, G_{2, k}, F_{1, k}, G_{1, k} $ are uniquely determined. Finding $F_{0, k}$ and $G_{0, k}$ can be done up to an additive constant. Adding [iii] and [jjj] leads to a closed form expression on $F_{0, k} + G_{0, k}$. Since the optimality condition does not depend on $F_0$ and $G_0$,  one may simply set $F_{0, k}$ to 0, and solve $G_{0, k}$ exactly. 
\qed

\section{The IBP algorithm}
\label{ss:ibp}

Computing the OT barycenter with the divergence $\ote^{\cU}$ can be shown to be equivalent to the KL projection problem:
\begin{equation}
\label{seq:ibp}
\tag{IBP}
\min_{\substack{\pi_1, \dots, \pi_K \\ \pi_k \in \cC_k \cap \cC'}} \sum_{k=1}^K w_k \widetilde{\kl}(\pi^k| \bK)\enspace,
\end{equation}
where $\cC_k = \{\pi \in \bbR^{n\times n}_+ | \pi \mathds 1 = \alpha_k\}$ and $\cC'= \{\pi \in \bbR^{n\times n}_+ | \exists \alpha \in \Delta_n, \enspace \pi_k^\top \mathds 1 = \alpha, \enspace \forall k = 1\dots K\}$.
The IBP algorithm amounts to performing iterative minimization on one constraint set $\cC$ at a time. Each step can be solved in closed form, leading to Sinkhorn-like iterations. By combining both iterations, one can write every iterate of the transport plan as $\pi^{(l)} =\diag(\ba^{(l)}) \bK \diag(\bb^{(l)})$ and perform the scaling operations on the variables $\ba, \bb$ given in algorithm \ref{salg:ibp}.

\begin{algorithm}
	\caption{IBP algorithm \cite{benamou14, chizat17}}
	\label{salg:ibp}
	\begin{algorithmic}
		\STATE {\bfseries Input:}  $ \alpha_1, \dots, \alpha_K$,  $\bK = e^{-\frac{\bC}{\varepsilon}}$
		\STATE {\bfseries Output:} $\alpha_{\ote^{\cU}}$ 
		\STATE Initialize all scalings $(b_k) $ to $\mathds 1$, 
		\REPEAT
		\FOR{$k=1$ {\bfseries to} $K$}
		\STATE $a_k \gets \left(\frac{\alpha_k}{\bK b_k}\right)$
		\ENDFOR
		\STATE $\alpha \gets  \prod_{k=1}^K ( \bK^\top a_k) ^{ w_k } $
		\FOR{$k=1$ {\bfseries to} $K$}
		\STATE $b_k \gets \left(\frac{\alpha}{\bK^\top a_k}\right)$
		\ENDFOR
		\UNTIL{convergence}
	\end{algorithmic}
\end{algorithm}
 
\section{Additional Proofs}

\label{ss:proof-alg1}

\paragraph{Proof of proposition \ref{prop:sdiv-kl}}
\begin{sprop}
	\label{sprop:sdiv-kl} Let $\alpha_1, \dots, \alpha_K \in \Delta_n$ and $\bK = e^{-\frac{\bC}{\varepsilon}}$. Let $\pi$ denote a sequence $\pi_1, \dots, \pi_K$ of transport plans in $\bbR_+^{n\times n}$ and the constraint sets $\cH_1=\{ \pi | \forall k, \, \pi_k \mathds 1 = \alpha_k\}$, and $\cH_2 = \{ \pi | \forall k\, \forall k', \, \pi_k^\top \mathds 1 = \pi_{k'} \mathds 1 \}$.
	The barycenter problem $\min_{\alpha \in \Delta_n} \sum_{k=1}^K w_k \sdiv(\alpha_k, \alpha)$ is equivalent to:
	\begin{align}
	\label{seq:sdiv-bar-kl}
	\begin{split}
	\min_{\substack{\pi \in \cH_1 \cap \cH_2 \\  d \in \bbR_+^n}} & \Bigg [ \varepsilon \sum_{k=1}^K w_k \kl(\pi_k| \bK \diag(d)) 
	 + \frac{\varepsilon}{2} \langle d - \mathds 1 , \bK (d - \mathds 1) \rangle \Bigg ] \enspace .
	\end{split}
	\end{align}
\end{sprop}
\proof
The barycenter problem of $\sdiv$ only depends on $\ote^{\cU}(\alpha, \beta) - \frac{1}{2}(\ote(\alpha, \alpha)$. Let's rewrite this expression using the IBP formulation and duality.
the IBP formulation \eqref{eq:otdef-uniform-kl} is explicitly given by: 
\begin{equation}
\label{seq:exact-ibp}
\ote^{\cU}(\alpha, \beta) = \min_{\substack{\pi \in \bbR^{n \times n}_+ \\ \pi \mathds 1 = \alpha, \pi^\top \mathds 1 = \beta}}  \varepsilon \widetilde{\kl}(\pi | \bK) - \varepsilon \sum_{i, j}\bK_{ij}
\end{equation}
And the autocorrelation term can be expressed via its dual problem:
\begin{align}
\label{seq:dual-auto-unif}
\ote^\cU(\alpha, \alpha) &= \max_{h\in \bbR^n} 2 \langle h, \alpha\rangle - \varepsilon \langle e^{\frac{h}{\varepsilon}}, \bK e^{\frac{h}{\varepsilon}}\rangle  - \varepsilon \sum_{i, j}\bK_{ij} \\
=&\max_{d\in \bbR_+^n} 2 \langle \varepsilon \log(d), \alpha\rangle - \varepsilon \langle d, \bK d\rangle - \varepsilon \sum_{i, j}\bK_{ij} \\
=&-\min_{d\in \bbR_+^n} - 2 \langle \varepsilon \log(d), \alpha\rangle + \varepsilon \langle d, \bK d\rangle + \varepsilon \sum_{i, j}\bK_{ij}
\end{align}
Moreover, on the constraint set $\cH_1 \cap \cH_2$, it holds $\alpha = \pi_k^{\top}$ for all $k$. Thus, denoting $\cH_2(\alpha) = \{ \pi | \forall k\, \forall k', \, \pi_k^\top \mathds 1 = \alpha\}$ the following can be written:

\begin{align*}
&\argmin_{\alpha \in \Delta_n} \sum_{k=1}^K w_k \sdiv(\alpha_k, \alpha) \\
&= \argmin_{\alpha \in \Delta_n} \min_{\pi \in \cH_1 \cap \cH_2(\alpha)} \sum_{k=1}^K w_k \varepsilon\widetilde{\kl}(\pi_k | \bK) + \min_{d\in \bbR_+^n} -  \langle \varepsilon \log(d), \alpha\rangle + \frac{1}{2}\varepsilon \langle d, \bK d\rangle - \frac{1}{2}\varepsilon \sum_{i, j}\bK_{ij} \\
&= \argmin_{\alpha \in \Delta_n} \min_{\substack{\pi \in \cH_1 \cap \cH_2(\alpha) \\  d \in \bbR_+^n}} \sum_{k=1}^K w_k \left( \varepsilon\widetilde{\kl}(\pi_k | \bK)  -  \langle \varepsilon \log(d), \alpha\rangle \right) + \frac{1}{2}\varepsilon \langle d, \bK d\rangle - \frac{1}{2}\varepsilon \sum_{i, j}\bK_{ij}\\
&= \argmin_{\alpha \in \Delta_n} \min_{\substack{\pi \in \cH_1 \cap \cH_2(\alpha) \\  d \in \bbR_+^n}} \sum_{k=1}^K w_k \left( \varepsilon \widetilde{\kl}(\pi_k | \bK)  -  \langle \varepsilon \log(d), \pi_k^{\top} \mathds 1\rangle \right) + \frac{1}{2}\varepsilon \langle d, \bK d\rangle - \frac{1}{2}\varepsilon \sum_{i, j}\bK_{ij} \\
&= \min_{\substack{\pi \in \cH_1 \cap \cH_2 \\  d \in \bbR_+^n}} \sum_{k=1}^K w_k \left( \varepsilon\widetilde{\kl}(\pi_k | \bK)  -  \langle \varepsilon \log(d), \pi_k^{\top} \mathds 1\rangle \right)+ \frac{1}{2}\varepsilon \langle d, \bK d\rangle - \frac{1}{2}\varepsilon \sum_{i, j}\bK_{ij} \\
&= \min_{\substack{\pi \in \cH_1 \cap \cH_2 \\  d \in \bbR_+^n}} \sum_{k=1}^K w_k \left(\varepsilon\widetilde{\kl}(\pi_k | \bK \diag(d))  - \varepsilon \langle \bK d, \mathds 1\rangle  + \varepsilon \sum_{ij}\bK_{ij} \right) + \frac{1}{2}\varepsilon \langle d, \bK d\rangle - \frac{1}{2}\varepsilon \sum_{i, j}\bK_{ij} \\
&= \min_{\substack{\pi \in \cH_1 \cap \cH_2 \\  d \in \bbR_+^n}} \sum_{k=1}^K \varepsilon w_k \widetilde{\kl}(\pi_k | \bK \diag(d)) - \varepsilon \langle \bK d, \mathds 1\rangle  + \frac{1}{2}\varepsilon \langle d, \bK d\rangle + \frac{1}{2}\varepsilon \sum_{i, j}\bK_{ij} \\
	&=\min_{\substack{\pi \in \cH_1 \cap \cH_2 \\  d \in \bbR_+^n}}  \sum_{k=1}^K \varepsilon w_k  \kl(\pi_k| \bK \diag(d)) 
+ \frac{\varepsilon}{2} \langle d - \mathds 1 , \bK (d - \mathds 1) \rangle \enspace .
\end{align*}
\qed

\section{Supplementary details on experiments}
\label{ss:expe}

\subsection{Barycenters of nested ellipses}
 We simulate each ellipse by generating random major and minor radii with a moving a center from the top left quarter corner to the bottom right quarter corner. The box constraints of the random generators of the radii are manually picked so that ellipses are more likely to be nested with an assymetric surrounding ellipse (see supplementary code). The full list of 10 images used to compute the barycenters is displayed in Figure \ref{sf:ellipses}. Each image has 60$\times$60 pixels. The ground OT cost function is the squared Euclidean cost over the unit square $[0,1]^2$. For entropy regularized distances (All except $W$), we set $\varepsilon$ to the lowest value guaranteeing no numerical instabilities in Sinkhorn's algorithm (this was particularly an issue for \emph{Sharp barycenters} $\alpha_{A_\varepsilon}$ of \citet{luise18}). Now we detail the algorithm used for each divergence $F$ defining each barycenter $\alpha_F$ of the experiment in Figure \ref{f:bar-ellipses}:
 
 \begin{enumerate}
 	\item $\ote^{\cU}$: OT with the uniform measure; computed using the IBP algorithm (Algorithm \ref{salg:ibp}).
  	\item $\sdiv$: Proposed debiased divergence; computed using the proposed algorithm (Algorithm \ref{alg:debiased-sinkhorn}).
  	\item $\ote^\otimes$: Computed using iterative IBP in minimization-majorization alternative algorithm. With \eqref{eq:reference-u}, one can linearize the concave negative $\kl$ penalty and solve the resulting problem using IBP iteratively and then update the $\kl$ term etc. This leads to a series of nested IBP loops.
  	\item $A_\varepsilon$: Sharp barycenters introduced by \citet{luise18}. Solved using accelerated gradient descent. This method required considerable manual effort to tune the learning rate in order to get an acceptable barycenter and was more prone to numerical instabilities.
  	\item Free support barycenters with $\sdiv$: introduced by \citet{luise19}, we used the online Python code provided by the authors which amounts to add or remove a dirac particle at each iteration and update their weights using Frank-Wolf's algorithm. The algorithm is stopped when no particules are created / removed.
  	\item $W$: non regularized Wasserstein distance. We used the accelerated interior point methods introduced by \citet{dongdong19} with the online matlab implementation provided by the authors.
 \end{enumerate}

\begin{figure}[!t]
\includegraphics[width=\linewidth]{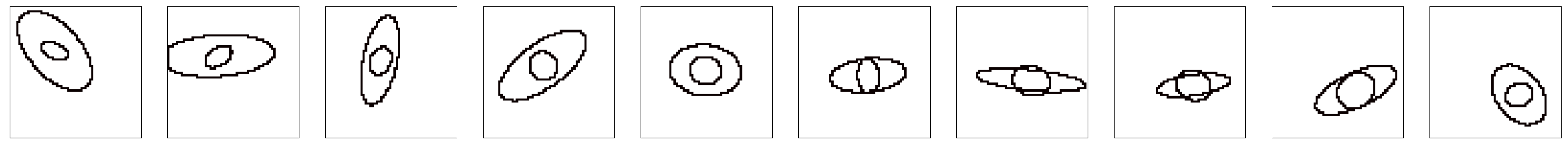}
\caption{All 10 nested ellipses images used to compute the barycenters of Figure \ref{f:bar-ellipses}. \label{sf:ellipses}}
\end{figure}

\subsection{Barycenters of 3D shapes}
The original 3D shapes (tore and rabbit) are taken from the \texttt{PyVista} \citep{pyvista} Python library. We preprocess the original meshes as follows. Each mesh is smoothed by 100 iterations of a Laplacian operator then the coordinates are centered and rescaled to fit within 95\% of the cube $(-1, 1)^3$. We sample 3D histograms of both meshes on a uniform
3D grid of size $200^3$. Both histograms are normalized and regularized by adding a $10^{-10}$ weight to avoid numerical errors. We set the lowest stable regularization  $\varepsilon = 0.01$ for the ground cost defined as the squared Euclidean distance over the $(-1, 1)^3$ cube. We compute weighted barycenters with the IBP algorithm \ref{salg:ibp} and the proposed debiased Sinkhorn barycenter algorithm \ref{alg:debiased-sinkhorn}. For each method, we use the weights $(w, 1-w)$ for $w \in [0, 0.25, 0.5, 0.75, 1.]$. The original meshes are shown in Figure \ref{sf:meshes}.

\begin{figure}[!t]
	\centering
	\includegraphics[width=0.3\linewidth]{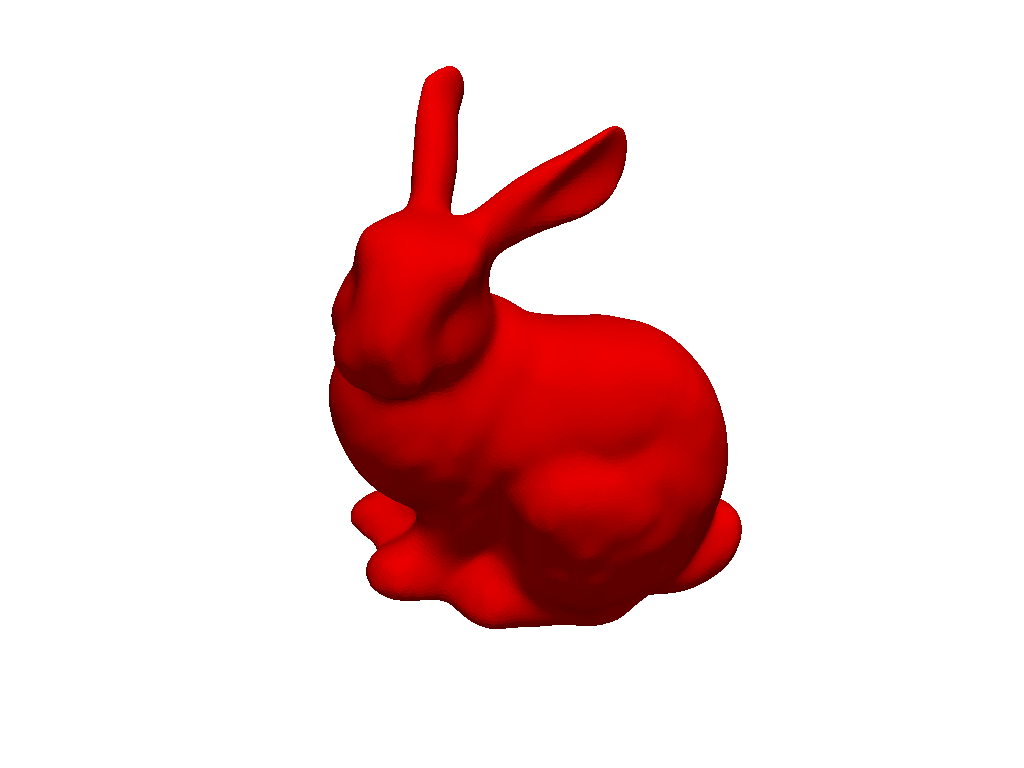}
	\includegraphics[width=0.3\linewidth]{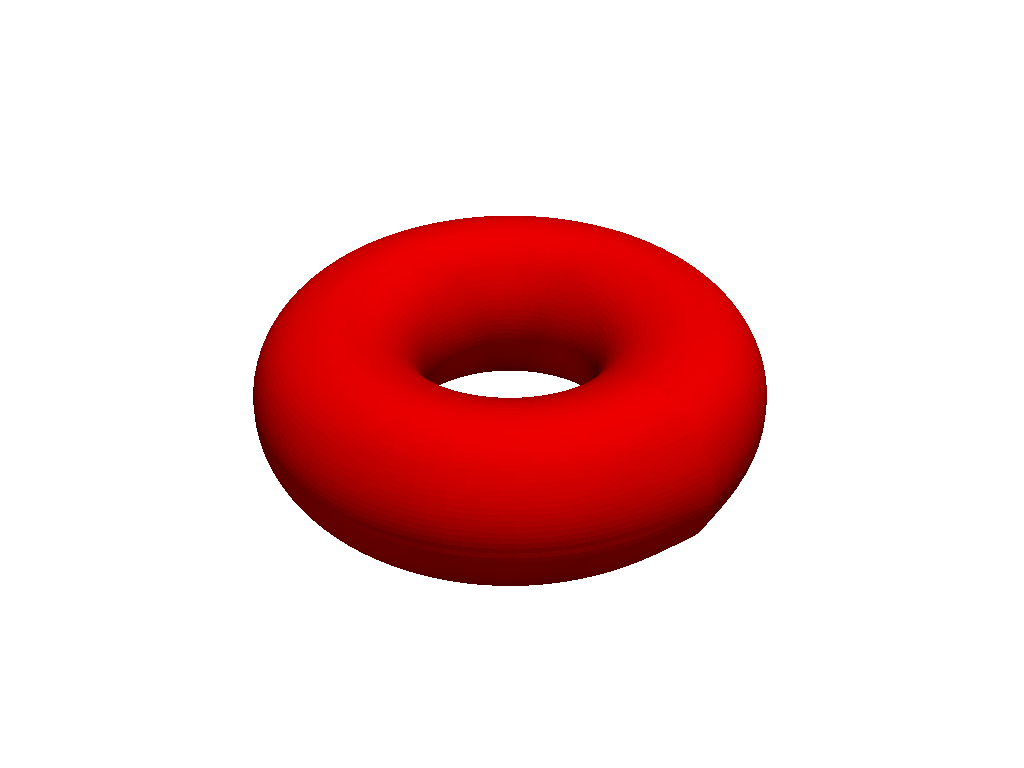}
	\caption{Input meshes used to compute the barycenters of 3D meshes. \label{sf:meshes}}
\end{figure}

\subsection{OT barycentric embeddings}
We use the Python library Torchvision that provides a fetch method to download the MNIST dataset. We first filter the data by keeping the labels (0, 1, 2, 3, 4) then select the first 500 samples. This constitutes the global dataset of the experiment. Then we randomly select $K = 50$ samples that will be considered as our learning dictionary $\cA$. Then for each sample (image) $\beta$ in the remaining 450 samples, we compute the weights $w \in \Sigma_K$ minimizing $\|\alpha_F(w) - \beta\|^2$ where $\alpha_F(w)$ is the weighted barycenter of the dictionary $\cA$. This leads to an embedding of 450 MNIST samples in a space of dimension $K$. We then use this embedding to train a Random Forest Classifier with 100 estimators using scikit-learn's default parameters (version 0.21.3).

}{}

\end{document}